\documentclass[runningheads]{llncs}
\usepackage[pdftex]{graphicx}
\usepackage{tikz}
\usepackage{comment}
\usepackage{amsmath,amssymb}
\usepackage{color}
\usepackage{subcaption}
\usepackage{booktabs}
\usepackage{multirow}
\usepackage[accsupp]{axessibility}

\pdfoutput=1

\begin{document}
\pagestyle{headings}
\mainmatter
\def\ECCVSubNumber{7450}

\title{VizWiz-FewShot: Locating Objects in Images Taken by People With Visual Impairments}

\titlerunning{VizWiz-FewShot: object localization}
\author{Yu-Yun Tseng$^*$ \and Alexander Bell$^*$ \and Danna Gurari \\
{\small * denotes equal contribution}}
\authorrunning{Y.-Y. Tseng et al.}
\institute{University of Colorado Boulder}

\maketitle

\begin{abstract}
We introduce a few-shot localization dataset originating from photographers who authentically were trying to learn about the visual content in the images they took.  It includes nearly 10,000 segmentations of 100 categories in over 4,500 images that were taken by people with visual impairments.  Compared to existing few-shot object detection and instance segmentation datasets, our dataset is the first to locate holes in objects (e.g., found in 12.3\% of our segmentations), it shows objects that occupy a much larger range of sizes relative to the images, and text is over five times more common in our objects (e.g., found in 22.4\% of our segmentations).  Analysis of three modern few-shot localization algorithms demonstrates that they generalize poorly to our new dataset.  The algorithms commonly struggle to locate objects with holes, very small and very large objects, and objects lacking text.  To encourage a larger community to work on these unsolved challenges, we publicly share our annotated few-shot dataset at \texttt{https://vizwiz.org}.
\end{abstract}

\section{Introduction}
\label{sec:introduction}
Our paper is motivated by the belief that people who are blind or with low vision (BLV) would benefit from the ability to locate objects in images that they take, whether with a bounding box or fine-grained segmentation.  For people with low vision, localization would enhance their use of magnification tools~\cite{afb_magnifiers,stangl2018browsewithme} by automatically enlarging the content of interest.  For all BLV users, they could have stronger privacy guarantees with services\footnote{Visual assistance services include Microsoft's Seeing AI, Google's Lookout, and TapTapSee. The popularity of such services is exemplified by companies' reports about hundreds of thousands of users and tens of millions of requests~\cite{desmond_microsofts_nodate,lee2020emerging,be_my_eyes_be_nodate}.} that describe their images if object localization algorithms were used in place of recognition algorithms.  That is because services could use localizations to obfuscate all content except the detected regions needed to justify predictions\footnote{Recorded evidence can be needed by companies for legal reasons.} and so remove accidentally captured private information in the background of images, which is a common occurrence for people with vision impairments~\cite{gurari2019vizwiz}.  Finally, automatic localization would also support users to independently edit their images, which is a feature some BLV photographers have requested.

Observing that BLV photographers take pictures showing a large number of objects (e.g., 16,400 nouns were used to describe less than 40,000 images taken by BLV photographers~\cite{dataset_VizWizCaption}), we are interested in the problem of few-shot learning.  Casting the problem as a few-shot learning problem means that developers can efficiently scale up the number of categories supported in order to locate the long-tail of categories.  That is because few-shot learning methods learn to locate a novel object category by observing only $K$ annotated examples, where $K$ is typically 1, 5, or 10 examples.  

\begin{figure}[t!]
\centering
	\includegraphics[width=\linewidth]{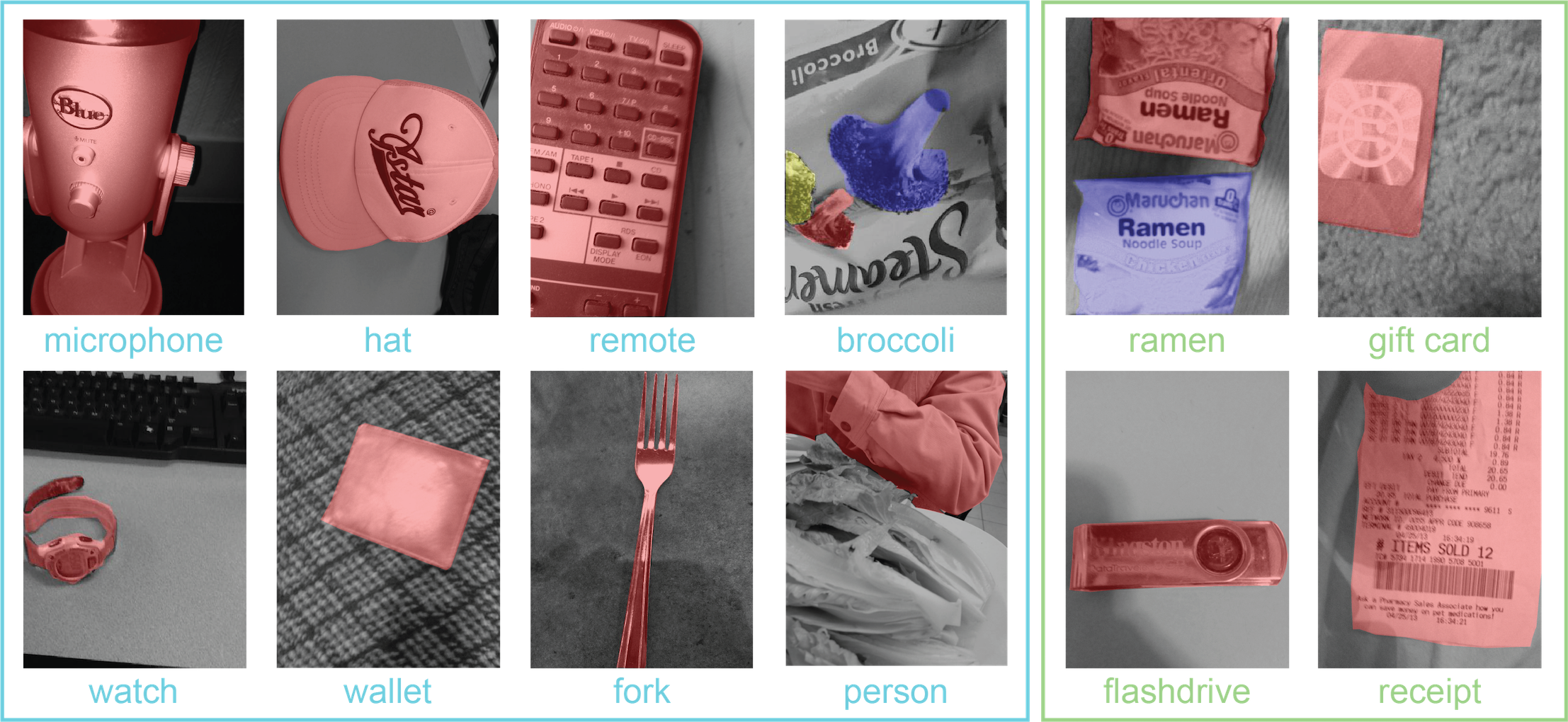} \hfill 
	\vspace{-2em}
  \caption{Examples from our VizWiz-FewShot dataset showing instance segmentation annotations we collected for images taken by people with vision impairments. Annotated categories span those that are in common with prior work (left green box) and unique to our dataset (right blue box).  These examples highlight novel aspects of our dataset, including that holes are permitted in our objects, objects vary considerably in how much of the image they occupy, and objects often feature text.  (Annotation overlay colors were selected based on the order the instance segmentations appear in our dataset, and so not object categories.)}
  \label{fig:motivation}
\end{figure}

To support our aim, we introduce a few-shot localization dataset that consists of 100 segmented categories in over 4,500 images taken by people with vision impairments.  The images were taken in authentic use cases where the photographers were soliciting human assistance to learn about their visual surroundings~\cite{bigham2010vizwiz}.  Examples of annotated images in our dataset are shown in Figure~\ref{fig:motivation}.  

We next analyze how our dataset compares to the four existing few-shot localization datasets~\cite{fsss_oneshot,fsss_FeatureWeighting,fsis_siameseMRCNN,fsod_attenRPN,dataset_FSS1000} to reveal both how our dataset is similar to and different from prior work.  We observe several unique aspects about our dataset.  First, it is the only dataset that indicates when and where holes are located in objects.  Holes are observed in 12.3\% of our instance segmentations.  Second, our dataset's objects exhibit a much larger range of sizes relative to the image sizes.  Finally, our dataset's objects contain text much more frequently.  Specifically, our analysis shows that 22.4\% of objects in our dataset contain text versus 4.6\% of the objects in the related instance sementation dataset, COCO-$20^{i}$~\cite{fsis_siameseMRCNN}.  We suspect the latter two unique aspects of our dataset stem from how images were curated.  While our images come from a real-world application where photographers were authentically trying to learn about their visual surroundings, existing datasets were contrived by scraping images from photo-sharing websites.  Altogether, we believe that our new dataset fills important gaps of existing datasets in the vision community by capturing a greater diversity of challenges that can arise in real-world applications.

We also benchmark top-performing few-shot learning object detection and instance segmentation algorithms on our new dataset.  We find that the algorithms perform poorly overall.  Our fine-grained analysis reveals that the algorithms commonly fail for objects that contain holes, very small and very large objects, and objects that lack text.

In summary, our contributions include: (1) a new few-shot localization dataset based on images that were taken in a real-world application, (2) the first few-shot localization dataset with metadata showing where holes are located in objects, (3) fine-grained analysis revealing unique aspects of our dataset compared to existing few-shot localization datasets, and (4) analysis of top-performing few-shot localization algorithms that reveals open algorithmic challenges for the vision community.  We expect this work will encourage the development of algorithms that can handle a greater diversity of challenges that arise in real-world applications.  We expect these advancements will, in turn, benefit a larger audience by facilitating the improvement of algorithms for application domains, such as robotics and wearable lifelogging, that face similar challenges including holes in objects, varying object sizes, and presence of text. 

\section{Related Work}
\label{sec:related}

\paragraph{Few-Shot Learning Datasets for Image Localization.}
Several dataset challenges have been proposed for few-shot object detection and few-shot instance segmentation: PASCAL-$5^{i}$ \cite{fsss_oneshot}, COCO-$20^{i}$ \cite{fsss_FeatureWeighting,fsis_siameseMRCNN}, ImageNet-LOC \cite{dataset_imagenet}, and FSOD Dataset \cite{fsod_attenRPN}. A limitation of existing datasets is that images come from contrived settings rather than authentic use cases where people are seeking to learn about their images.  Specifically, images were curated by scraping images from the Internet that were tagged with pre-defined categories of interest.  To our knowledge, we are introducing the first few-shot dataset challenges based on images that originate from authentic use cases where people took pictures to learn about the content.  Our dataset offers new categories that are applicable to real-world applications.  In addition, our dataset provides metadata showing holes in objects, which is a unique feature that creates new challenges toward few-shot problems.  Finally, it provides additional real-world challenges such as a larger range of object sizes and a higher prevalence of objects containing text.
 
\vspace{-0.5em}\paragraph{Few-Shot Algorithms for Image Localization.}
Few-shot learning was introduced to the community in 2017 for object detection~\cite{FSOD_fewExample} and in 2018 for instance segmentation~\cite{fsis_siameseMRCNN}.  Since, a large number of algorithms have been proposed that largely are based on two types of approaches: meta-learning and fine-tuning.  To assess how state-of-the-art methods perform on our new dataset, we benchmark the top-performing few-shot  object detection and instance segmentation algorithms for which code is publicly-available. Overall, we observe poor performance from these algorithms~\cite{fsod_DeFRCN,fsis_siameseMRCNN,fsis_FAPIS,fsis_yolact}. From our fine-grained analysis, we find this dataset is challenging for algorithms due to the presence of holes in objects, very small objects, very large objects, and objects that lack text.

\vspace{-0.5em}\paragraph{Datasets Originating from People With Vision Impairments.}
In recent years, a growing number of publicly-available datasets have been proposed to facilitate the development of algorithms that can work well on images taken by people with vision impairments~\cite{bhattacharya2019does,chen2022grounding,chiu2020assessing,dataset_VizWiz,dataset_VizWizCaption,dataset_KVAQ,dataset_orbit,gurari2018predicting,gurari2019vizwiz,zeng2020vision}. For example, existing datasets support the development of algorithms for predicting answers to visual questions~\cite{dataset_VizWiz,dataset_KVAQ}, recognizing objects in videos~\cite{dataset_orbit}, and describing images with captions~\cite{dataset_VizWizCaption}.  Complementing prior work, we introduce a dataset for localizing objects in images taken by BLV photographers, either using a bounding box or segmentation. We expect success with developing localization algorithms for images taken by BLV photographers to directly benefit BLV photographers and to, more generally, support a larger number of real-world applications that encounter similar visual characteristics found in our dataset, such as robotics and wearable lifelogging applications.

\section{VizWiz-FewShot Dataset}
\label{sec:dataset}

We introduce a dataset that we call ``VizWiz-FewShot".  It consists of localization annotations for images taken by people with vision impairments who authentically were trying to learn about their visual surroundings.

\subsection{Dataset Creation}

\paragraph{Data Source.}
Our dataset extends the VizWiz-Captions dataset~\cite{dataset_VizWizCaption}, which consists of images taken by people with vision impairments paired with five crowdsourced captions per image.  The photographers took and shared these images in order to solicit assistance from remote humans in recognizing the contents in the images~\cite{bigham2010vizwiz}. We leverage the data in both the train and validation splits, which offers a starting point of 31,181 captioned images.  

\vspace{-0.5em}\paragraph{Category Selection.}
We chose $100$ categories to locate in the images.  These categories both support backward compatibility with popular few-shot localization datasets and reflect important categories for people with vision impairments.  To select the categories, we first quantified the frequency of all nouns that appeared in at least two of the five captions per image for our images.  We then selected $72$ non-ambiguous categories that overlap with four existing few-shot localization datasets: MS COCO \cite{dataset_coco}, PASCAL VOC \cite{dataset_voc}, FSS-1000 \cite{dataset_FSS1000}, and FSOD \cite{fsod_attenRPN}.  We also selected $28$ non-ambiguous categories that are unique to our target population, by choosing categories that refer to physical objects. All 100 selected categories have at least 10 examples. 
 
\vspace{-0.5em}\paragraph{Data Filtering.}
We next filtered the images to only retain those that contained at least one of our 100 categories.  First, we removed images which did not mention any of our categories within at least two of their respective captions.  Then, the authors subsequently verified that each remaining image contained at least one object that fit the precise definition of at least one of our categories. For example, our automatic collection of images with the category ``pen" retrieved some images of pencils without pens and so we removed those images. After filtering, we had total of $4,930$ images.

\vspace{-0.5em}\paragraph{Annotation Tasks.}
After iterative prototyping, we settled on a workflow similar to prior work~\cite{dataset_coco}, such that we first used an image classification task to flag which categories of interest are present in each image and then an instance segmentation task to locate every instance of each category.  For both tasks, we utilized templates provided by Amazon Mechanical Turk (AMT).

For image classification, crowdworkers were shown an image and asked to select all categories that were present, if any.  Since showing all $100$ categories at the same time could overwhelm crowdworkers and ultimately lead to lower quality results, we instead showed a subset of categories at a time (i.e., $\sim$20).

For instance segmentation, crowdworkers were shown an image with the list of categories known to be present from the image classification task and asked to locate each instance of every category. Like prior work, our annotation tool supported users to create a series of clicks to generate polygons.  Going beyond prior work, in addition to being able to draw  `positive' polygons to locate object boundaries, our tool also enabled users to create `negative' polygons in order to capture when objects contained holes. We offered extensive instructions with our task to cover edge-case scenarios, including how to annotate the presence of holes and how to handle occlusions.  

\vspace{-0.5em}\paragraph{Annotation Collection.}
We implemented several quality control methods to support our collection of high-quality annotations. First, we only accepted workers who already had completed at least 500 AMT tasks with at least a 99\% approval rating. For the more complex instance segmentation task, we also required workers to successfully pass a qualification test consisting of nine challenging annotation edge cases (described in the Supplementary Materials).  We then collected redundant results from multiple unique workers for both tasks.  For image classification, we collected three results per image and flagged a category as present if at least one worker indicated so.  For instance segmentation, we collected two sets of annotations per image-category pair and then computed intersection over union ($IoU$) scores to determine how to establish a ground truth segmentation per image. When $IoU \geq 0.8$, we randomly chose one of the annotations as the ground truth.  Otherwise, the authors reviewed the pair of annotations to choose one as the ground truth (or, in exceptional cases, discarded both annotations).  Finally, we paid above minimum wage to better incentivize the workers.\footnote{Average hourly wage was \$$8.00$ and \$$9.61$ for classification and IS respectively.}  Upon completion, we had a total of 9,861 segmented objects in 4,622 images.  

\subsection{Dataset Analysis}

We now analyze the VizWiz-FewShot dataset and compare it to the other mainstream few-shot localization datasets.

\vspace{-0.5em}\paragraph{VizWiz-FewShot-IS (Instance Segmentation).}
We first characterize our few-shot instance segmentation dataset and compare it to the only other few-shot instance segmentation dataset we are aware of: COCO-20\textsuperscript{i} \cite{dataset_coco}, which has a total of 80 categories.  We compute for every instance segmentation the following metrics:
\begin{itemize}
    \item \textbf{Mass center}: location of the center of mass pixel for each object relative to the image coordinates.  Consequently, an object's x-coordinate and y-coordinate values can range from 0 to 1.
    \item \textbf{Boundary complexity}: ratio of the area of an instance to the length of its perimeter, also known as isoperimetric inequality.  Values range from 0 to 1, with lower values representing more complex boundaries.
    \item \textbf{Image coverage}: percentage of pixels each instance segmentation occupies from the entire image.
    \item \textbf{Prevalence of text}: flag indicating if Microsoft Azure's optical character recognition (OCR) API returned text for an image, after masking out all content except for the instance segmentation.
    \item \textbf{Prevalence of holes}: flag indicating if any holes are present paired with the percentage of pixels each hole occupies from the instance segmentation when any holes are present.
\end{itemize}
In what follows, we report the statistics summarizing the results for all instance segmentations for each dataset.\footnote{For efficiency, we evaluated the presence of text for a random sample of images in COCO-20\textsuperscript{i} that is comparable to the number of images in our dataset: 8,000.}  

\begin{figure}[t!]
\includegraphics[width=0.97\textwidth]{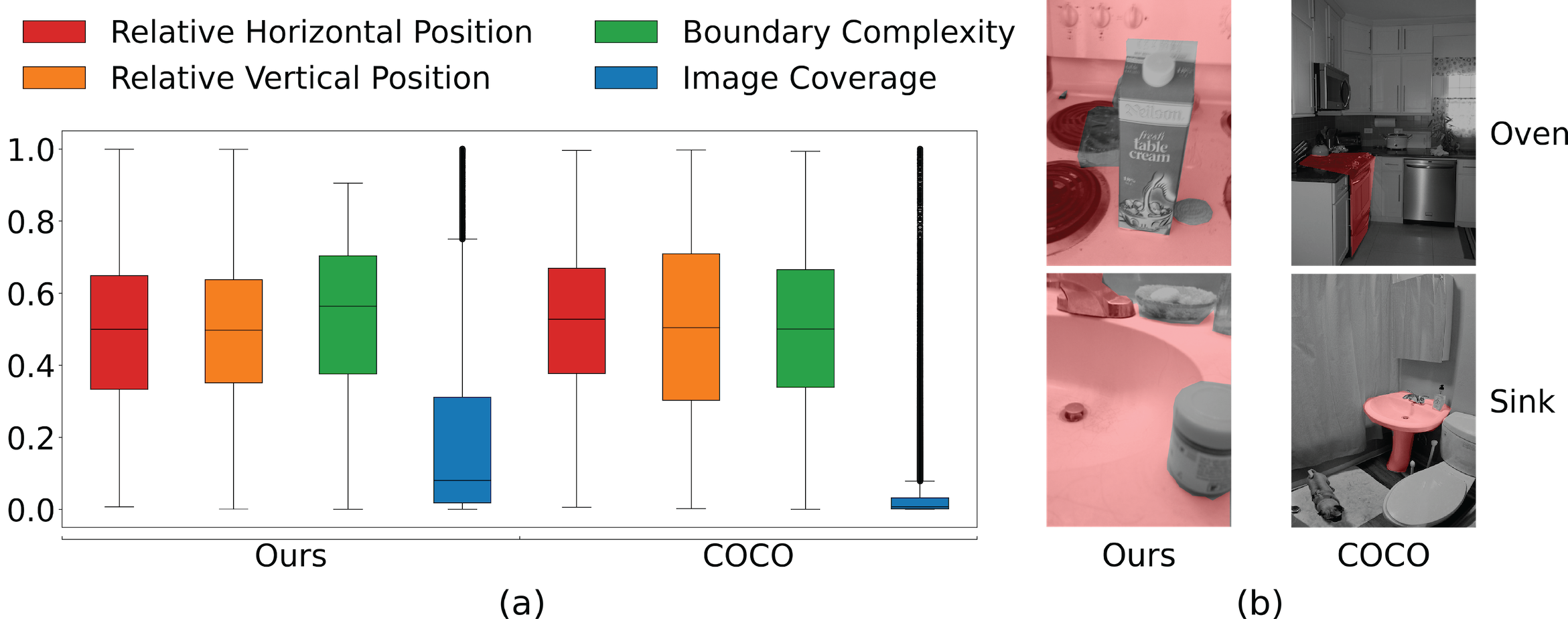}
\vspace{-0.7em}
\caption{Comparison of our dataset with the only existing few-shot instance segmentation dataset, COCO-20\textsuperscript{i}. (a) Summary statistics for all segmented objects in each dataset are shown in the box plot with respect to the location (i.e., relative x-coordinates and y-coordinates for mass center), boundary complexity, and image coverage.  The box plot's central mark denotes the median score, box edges the 25th and 75th percentiles scores, whiskers the most extreme data points not considered outliers, and individually plotted points the outliers. (b) Annotations from both datasets exemplify our quantitative finding that an object with larger image coverage is an outlier in COCO-20\textsuperscript{i} while common in our dataset.}
\label{fig:instance-seg-comparions}
\end{figure}

\begin{figure}[t!]
\centering
	\includegraphics[width=0.99\linewidth]{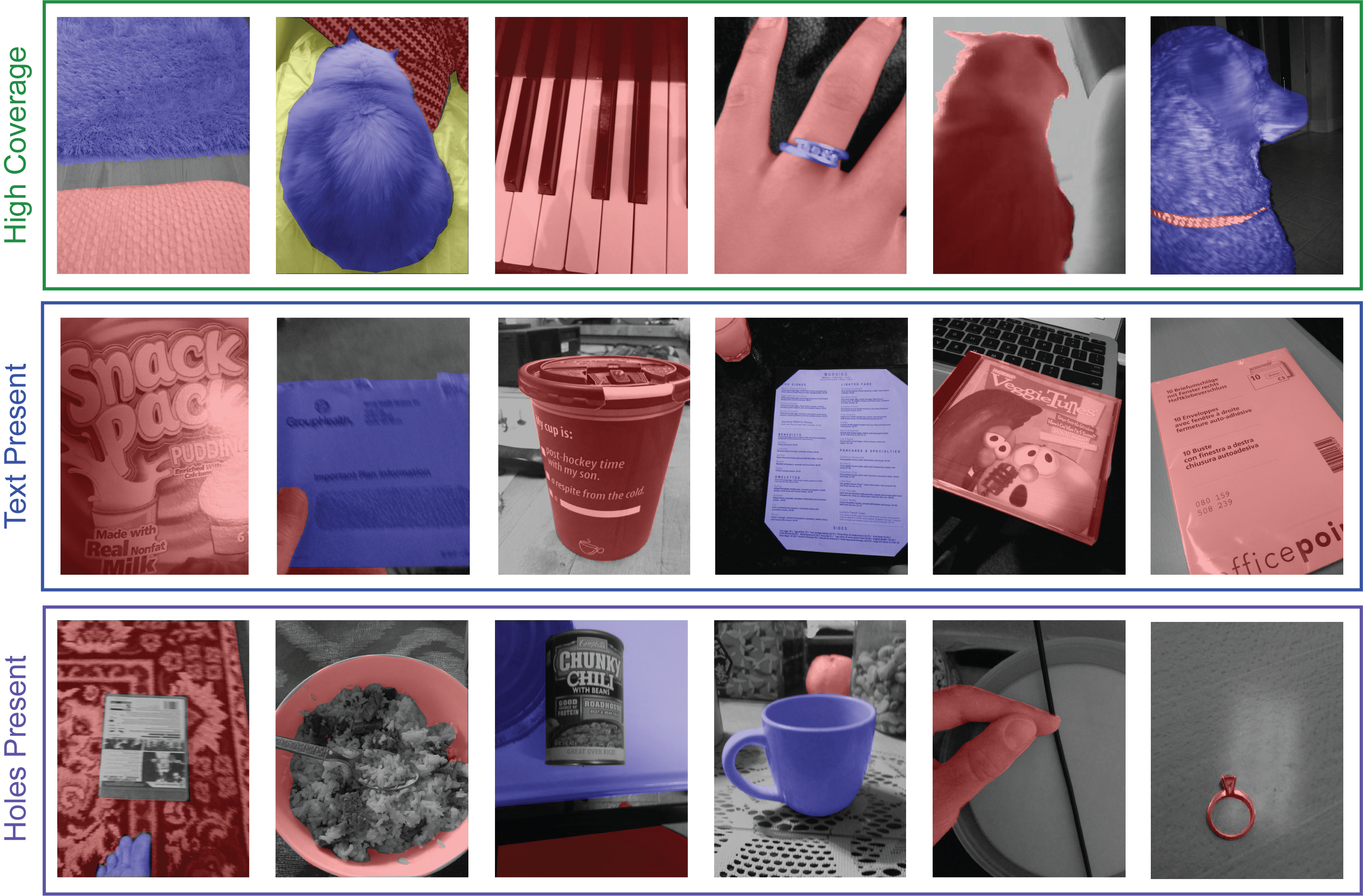} \hfill 
  \vspace{-0.75em}
  \caption{Examples from our VizWiz-FewShot dataset illustrating its unique aspects, specifically the high variability of object size relative to images, high prevalence of text in objects, and inclusion of holes in segmentations.}
  \label{fig:vizwiz-is-unique-aspects}
\end{figure}

Results for \emph{boundary complexity}, object location (i.e., \emph{mass center}), and \emph{image coverage} are shown in Figure~\ref{fig:instance-seg-comparions}(a).  Amongst these metrics, the only major difference between the two datasets is \emph{image coverage}. For example, objects in our dataset represent on average roughly six times more relative area in images than those in COCO-20\textsuperscript{i}.  We exemplify this finding qualitatively by showing in Figure~\ref{fig:instance-seg-comparions}(b) how annotations of two types of content, ``sink" and ``oven", dramatically differ in image coverage across the two datasets.  We attribute the prevalence of larger relative object sizes in our dataset to the fact that photographers in an authentic use case where they are trying to learn about content take up-close pictures of the content.  Another key distinction about our dataset is that we observe a considerably larger variability for the image coverage in our dataset.  Qualitative results in Figure~\ref{fig:vizwiz-is-unique-aspects} exemplify this range of relative area occupied by segmentations in our dataset. This finding highlights that a benefit of our dataset is that it encourages the design of algorithms that will be able to handle a larger range of relative object sizes in images.  

\begin{figure}[b!]
\centering
	\includegraphics[width=1.0\textwidth]{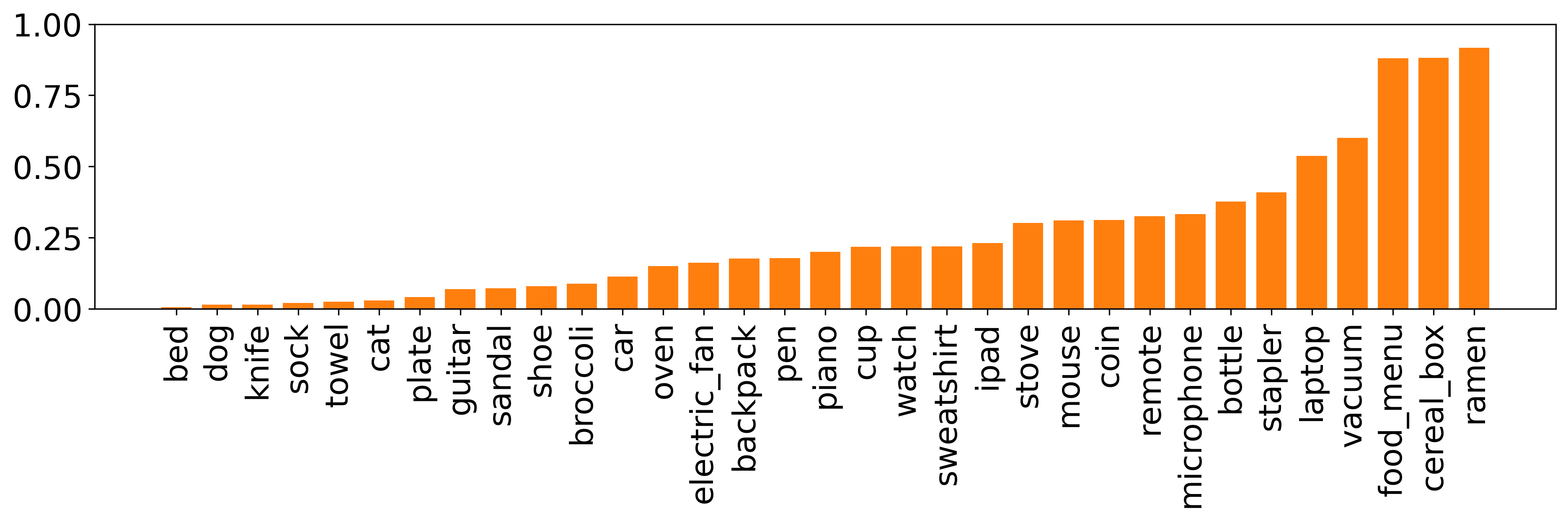} \hfill 
  \vspace{-2em}
  \caption{Proportion of instances with text on a per-category basis for each third category in our dataset, sorted them by frequency of text.}
  \label{fig:text-stack-histogram}
\end{figure}

\begin{figure}[b!]
\centering
	\includegraphics[width=1.0\textwidth]{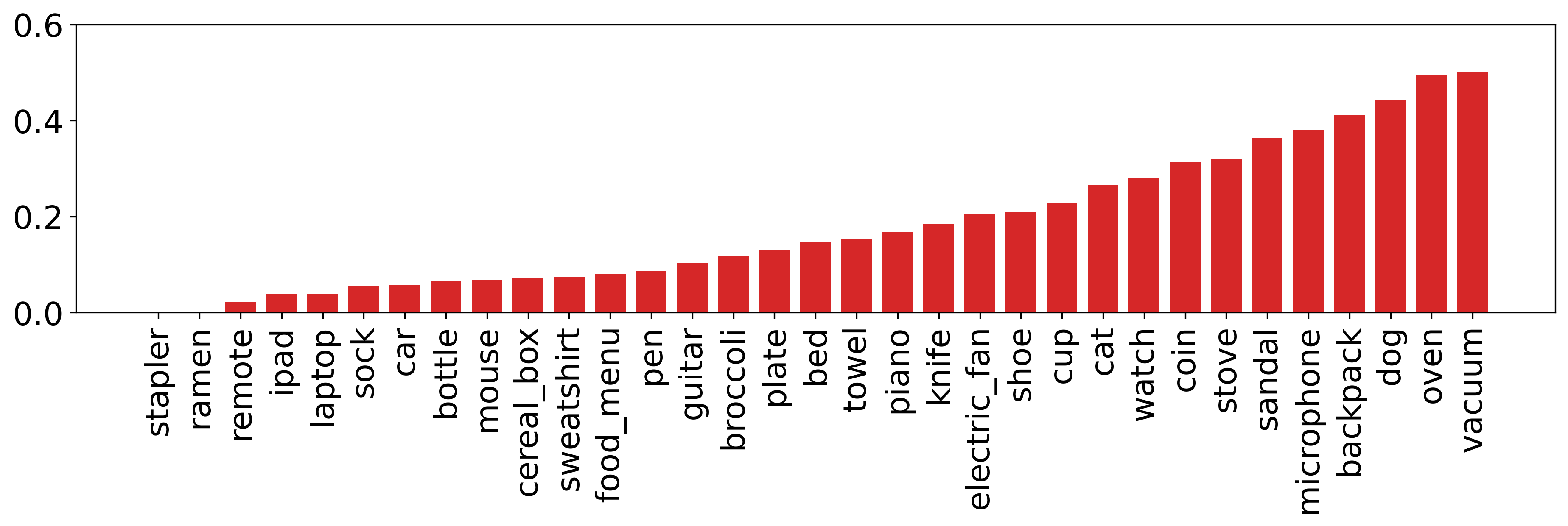} \hfill 
  \vspace{-2em}
  \caption{Proportion of instances with holes on a per-category basis for each third category in our dataset, sorted by frequency of holes.}
  \label{fig:hole-stack-histogram}
\end{figure}

When analyzing the \emph{prevalence of text}, we find that 22.4\% of instances in our dataset include text compared to only 4.6\% in COCO-20\textsuperscript{i}.  We show examples of objects in our dataset that contain text in Figure~\ref{fig:vizwiz-is-unique-aspects}, including of cups, menus, cereal boxes, and albums.  We also show in Figure~\ref{fig:text-stack-histogram} the frequency at which text is found in a sample of our categories.  Categories that more commonly contain text include ramen, food menu, packet, and gift card.  Categories that rarely contain text include dog, vase, house, and spoon.  We hypothesize from our findings that algorithm developers working on COCO-20\textsuperscript{i} may have a bias to disregard text.  We expect our dataset will inspire developers to consider how to take advantage of text recognition methods as potential predictive cues for locating objects with few-shot localization algorithms.

Finally, a unique feature of our dataset that is not supported in COCO-20\textsuperscript{i} is locating the \emph{holes} in objects.  We define a hole as any area in an object that does not belong to the object itself since our goal is to locate all pixels belonging to each category of interest. Thus, a hole may manifest as a property of an object itself (i.e. a ring), an object's orientation (i.e. a side view of an open armrest on a chair), or an occlusion on the object (i.e. a plate partially occluded by food).  In total, 12.3\% of the instances in our VizWiz-FewShot-IS contain holes.   As shown in Figure~\ref{fig:hole-stack-histogram}, some of the object categories with the highest proportion of instance segmentations that contain holes are chairs, sandals, bracelets, and bowls.  For instance, 21.1\% of the bowl instances have holes, likely because bowls typically contain food in them.  We attribute the high frequency of holes to two causes.  First, is that the objects intrinsically contain them; e.g., chairs, stools, and sandals.  The second reason is that large-appearing objects get occluded by foreground objects, such as occlusions on rugs, bowls, and plates.  Corroborating this hypothesis, we find that the percentage of instances with holes increases with object size, suggesting that larger objects tend to have hole-type occlusions more frequently than smaller objects (results shown in the Supplementary Materials).  We also observe that certain categories regularly have a larger percentage of hole pixels in them, such as bowls which typically are occluded by a large amount of food (results are shown in the Supplementary Materials).  We anticipate the need to recognize holes will increase our dataset's difficulty for computer vision models since they will need to go beyond merely locating the outermost boundary of objects to also understanding which interior pixels should belong to the objects.

\begin{figure}[b!]
    \centering
    \includegraphics[width=1.0\textwidth]{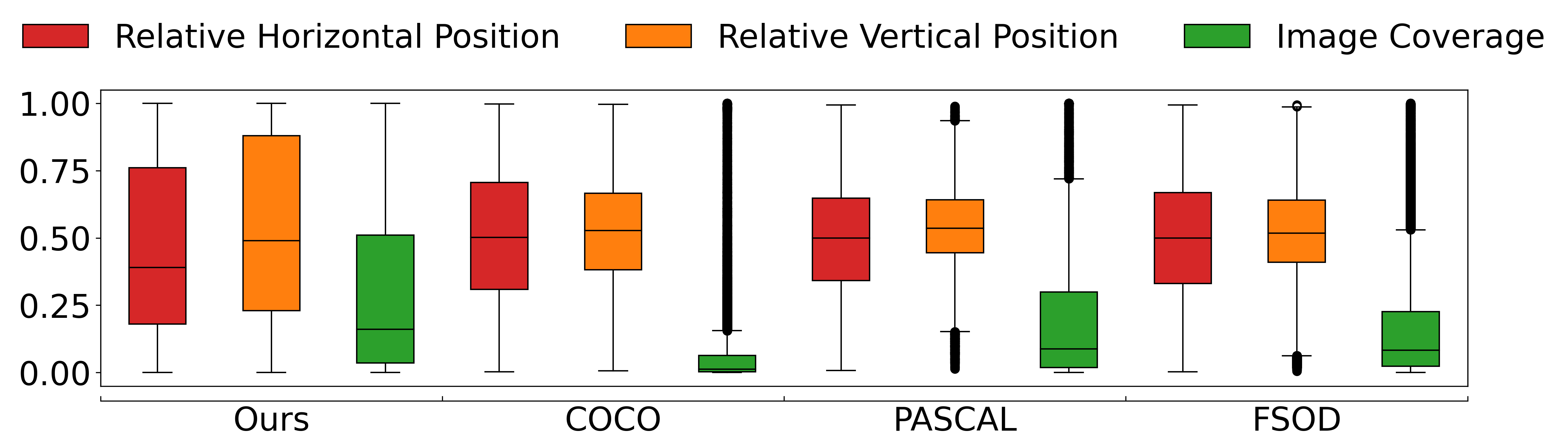}
    \vspace{-1.5em}
    \caption{Box plot showing how objects in our dataset compare to those in the three existing few-shot object detection datasets with respect to relative position and image coverage.}
     \label{fig:od-analysis}
\end{figure}

\vspace{-0.5em}\paragraph{VizWiz-FewShot-OD (Object Detection).}
We next characterize our dataset in the object detection setting and how it compares to the three mainstream few-shot object detection datasets: COCO-20\textsuperscript{i}~\cite{fsss_FeatureWeighting}, PASCAL-5\textsuperscript{i}~\cite{fsss_oneshot}, and FSOD~\cite{fsod_attenRPN}.\footnote{We use both the train and validation splits from each of the mainstream datasets for analysis. We randomly sample 10\% of the annotations from COCO-20\textsuperscript{i} due to its large size, and we use all annotations from PASCAL-5\textsuperscript{i} and FSOD.}  To support comparison, we convert each instance segmentation in our dataset into its bounding box representation. For every dataset, we compute for each object detection its relative position and image coverage.\footnote{We exclude from consideration the other three metrics used to analyze the instance segmentations because boundary complexity is no longer relevant, text prevalence could be incorrect due to the bounding box extending beyond an object's boundaries, and none of the other datasets located holes in objects.}  Summary statistics for each dataset are shown in Figure~\ref{fig:od-analysis}.

One key distinction of our dataset is the greater variability in the \emph{relative positions} of its objects.  This finding contrasts a common photographer's bias of beautifully capturing the contents of interest near the center of images.  We suspect this greater diversity of object positions stems from the inability of BLV photographers to inspect the images to guarantee that they are centering the contents of interest in their images and their inability to verify that clutter gets excluded from the background of their images.  

Another distinction in our dataset is its bias towards having objects positioned on the left side of images, as exemplified by the mean and median relative horizontal position being 0.45 and 0.39 respectively.  One possible reason for this bias may be a commonality in how the photographers take images.  Specifically, when a person is trying to learn about a particular object often the person holds the content of interest in the left hand while taking a picture of it with the right hand.  This scenario assumes a tendency in society for people to be right-handed.

Finally, we observe that bounding boxes in our dataset tend to cover more of an image than two of the four existing datasets: COCO-20\textsuperscript{i} and FSOD.  Image coverage of objects in our dataset is comparable to PASCAL, which we attribute to PASCAL's focus on iconic images with salient objects \cite{dataset_coco} and our dataset's inclusion of images of objects taken up-close for visual assistance.

\section{Algorithm Benchmarking}
\label{sec:benchmark}
We now present our results from benchmarking top-performing computer vision algorithms on our VizWiz-FewShot dataset.

To support use of our annotated data for few-shot localization tasks,
we create a 4-fold cross-validation format and split our 100 object categories into four sets, where $i={0,1,2,3}$ for the $i^{th}$ fold.  This approach mimics the settings used for the few-shot datasets PASCAL-$5^{i}$~\cite{fsss_oneshot} and COCO-$20^{i}$~\cite{fsss_FeatureWeighting,fsis_siameseMRCNN}.  We refer to the resulting datasets for few-shot instance segmentation and few-shot object detection as VizWiz-FewShot-IS-${25}^{i}$ and VizWiz-FewShot-OD-${25}^{i}$ respectively. 

We evaluate the trained models using $mAP$ and $mAP_{50}$. $mAP$ originates from the MS COCO object detection challenge~\cite{dataset_coco} and is frequently used to evaluate algorithms for FSIS~\cite{fsis_siameseMRCNN,fsod_metarcnn,fsis_FGN,fsis_FAPIS} and FSOD~\cite{fsod_review1}. $mAP$ refers to the mean of Average Precision ($AP$) for all categories and is an average across the IoU threshold of $0.5:0.05:0.95$ for ground truth and prediction regions.  The only difference betweent FSOD and FSID is that that former is evaluated based on bounding boxes while the latter is evaluated based on mask areas. We also present results with respect to $mAP_{50}$, where only threshold $0.5$ is used, since this approach facilitates the comparison with datasets such as Pascal VOC.  Our evaluation is based on when $K={1,3,5,10}$ shots are available.

\subsection{Few-shot Instance Segmentation Algorithms}
\label{subsec:algo_FSIS}
We benchmarked the top-performing FSIS algorithm for which code is publicly-available and can be successfully deployed on modern GPUs\footnote{We discuss the limitations of other FSIS algorithms for benchmarking on our dataset in the Supplementary Materials.}.  Specifically, we evaluated the algorithm YOLACT~\cite{fsis_yolact}, which was originally proposed outside of a few-shot setting, and then was subsequently shown to yield strong results on 
COCO-20\textsuperscript{i} when fine-tuned for FSIS~\cite{fsis_FAPIS}.  When using the codebase as is on our new FSIS dataset, the performance on novel classes is consistently negligible (i.e., $mAP$ around 0).  We found this occurs because the default hyperparameters leads to training loss explosion. Consequently, we tested with different hyperparameters.  Specifically, we (1) explored four learning rates in decreasing order from the original setting (i.e., 1e-3) to a value where saw convergence (i.e., 2e-5), (2) explored weights for the bounding box loss and mask loss in increasing order from 0 to 15 with an increment size of 1, (3) resized all images to match MS COCO's resolutions (i.e., $640 \times 480$), and (4) removed object instances of which the areas exceed that of MS COCO (i.e., instances are filtered based on the size range in MS COCO).  

\vspace{-0.5em}\paragraph{Overall performance:}
Results are shown in Table~\ref{table:overall_performance}. We report results with respect to each fold as well as the mean across all folds. 

Overall, the model performs poorly on VizWiz-FewShot-IS-${25}^{i}$.  Moreover, the performance is much worse on our dataset than observed on the original dataset for which it was proposed~\cite{fsis_FAPIS}; e.g., $mAP_{50}$ score of $2.48$ compared to $17.1$ for $1$-shot and $5.17$ compared to $18.9$ for $5$-shot  for VizWiz-FewShot-IS-${25}^{i}$ and COCO-20\textsuperscript{i} respectively. These findings motivate the benefit of our dataset in providing a challenging problem for the vision community. 

\begin{table*}[b!]
  \centering
  \caption{Overall performance of the few-shot algorithms on our VizWiz-FewShot dataset presented in 4-fold validation style. The FSIS algorithm is benchmarked on VizWiz-FewShot-IS-$25^{i}$, and the FSOD algorithms are benchmarked on VizWiz-FewShot-OD-$25^{i}$.}
    \begin{tabular}{lllrrrrrrrrrr}
    \toprule

& & & \multicolumn{2}{c}{$25^{0}$} & \multicolumn{2}{c}{$25^{1}$} & \multicolumn{2}{c}{$25^{2}$} & \multicolumn{2}{c}{$25^{3}$} & \multicolumn{2}{c}{mean} \\
~ & ~~~~~~ & shots & mAP & mAP\textsubscript{50} & mAP & mAP\textsubscript{50} & mAP & mAP\textsubscript{50} & mAP & mAP\textsubscript{50} & mAP & mAP\textsubscript{50} \\

\midrule \midrule
\multirow{4}{*}{\rotatebox[origin=c]{90}{FSIS}} &
\multirow{4}{*}{\rotatebox[origin=c]{90}{YOLACT}}
& $k=1$ & 1.87 & 2.5 & 2.91 & 3.51 & 1.39 & 1.79 & 1.08 & 2.13 & 1.81 & 2.48 \\
& & $k=3$ & 2.31 & 2.81 & 4.48 & 5.24 & 2.35 & 2.78 & 3.59 & 4.52 & 3.18 & 3.84 \\
& & $k=5$ & 3.45 & 4.30 & 4.84 & 5.67 & 4.34 & 5.14 & 4.39 & 5.56 & 4.25 & 5.17 \\
& & $k=10$ & 5.97 & 7.69 & 7.71 & 9.02 & 6.18 & 7.18 & 5.82 & 7.38 & 6.42 & 7.82 \\

\midrule

\multirow{4}{*}{\rotatebox[origin=c]{90}{FSOD}} &
\multirow{4}{*}{\rotatebox[origin=c]{90}{DeFRCN}}
& $k=1$ & 3.45 & 5.80 & 4.67 & 8.33 & 3.51 & 5.10 & 4.51 & 8.19 & 4.03 & 6.85 \\
& & $k=3$ & 6.80 & 11.65 & 7.81 & 13.85 & 7.26 & 11.74 & 7.88 & 14.05 & 7.43 & 12.82 \\
& & $k=5$ & 8.99 & 15.19 & 11.26 & 19.13 & 10.60 & 16.95 & 11.23 & 19.11 & 10.52 & 17.60 \\
& & $k=10$ & 11.24 & 21.34 & 13.36 & 25.68 & 11.94 & 22.07 & 13.91 & 24.76 & 12.61 & 23.46 \\

\midrule

\multirow{4}{*}{\rotatebox[origin=c]{90}{FSOD}} &
\multirow{4}{*}{\rotatebox[origin=c]{90}{YOLACT}}
& $k=1$ & 2.05 & 2.61 & 2.84 & 3.66 & 1.61 & 1.97 & 1.91 & 2.26 & 2.10 & 2.63 \\
& & $k=3$ & 2.45 & 3.05 & 4.41 & 5.53 & 2.58 & 3.22 & 3.94 & 4.89 & 3.35 & 4.17 \\
& & $k=5$ & 3.46 & 4.44 & 4.87 & 5.88 & 4.82 & 5.68 & 4.72 & 5.81 & 4.47 & 5.45 \\
& & $k=10$ & 6.27 & 7.89 & 7.60 & 9.29 & 6.61 & 7.90 & 6.06 & 7.86 & 6.64 & 8.24 \\

    \bottomrule
    \end{tabular}%
  \label{table:overall_performance}%
\end{table*}%

\vspace{-0.5em}\paragraph{Fine-grained analysis:}
To identify what make the dataset difficult, we next analyze the model's performances with respect to (1) image quality, (2) object size, and (3) presence of text. To do so, we distribute the test examples into subsets with respect to each of the following factors:

\begin{itemize}
    \item \textbf{Image quality}: Leveraging metadata from prior work~\cite{dataset_VizWizCaption} which indicates how many from five crowdworkers indicated an image is insufficient quality to recognize the content, we classify an image as ``high quality" when none indicate insufficient quality and ``medium quality" when one or two crowdworkers flagged the image as insufficient quality.  We exclude even lower quality images from our analysis since these are rare in our test set.
    % Definition: https://arxiv.org/pdf/2002.08565.pdf
    \item \textbf{Object size}: The target object size is calculated based on the number of pixels in the instance segmentations. We divide the dataset into small, medium, and large sizes, such that the numbers of images in each set are evenly distributed. This resulted in the following thresholds: ${350}^{2}$ and ${900}^{2}$.
    \item \textbf{Presence of text}: We used the metadata collected for Section~\ref{sec:dataset} that determined whether an object has text on it using OCR on background-masked instance images to flag whether text is present.
    \item \textbf{Presence of hole(s)}: We used additional metadata from Section~\ref{sec:dataset}, indicating if each instance segmentation contains a hole, to flag if a hole is present.
\end{itemize}

\begin{table*}[b!]
  \centering
  \caption{Fine-grained analysis on the performance of FSIS and FSOD models on VizWiz-FewShot presented in $mAP$.}
    \begin{tabular}{cllrrrrrrrrr}
    \toprule
    
& & & \multicolumn{2}{c}{Image Quality} & \multicolumn{3}{c}{Object Size} & \multicolumn{2}{c}{Presence of Text} & \multicolumn{2}{c}{Presence of Holes}\\
~ & ~~~~ & shots & Medium & High & Small & Medium & Large & Yes & No & Yes & No \\
\midrule \midrule
\multirow{4}{*}{\rotatebox[origin=c]{90}{FSIS}} &
\multirow{4}{*}{\rotatebox[origin=c]{90}{YOLACT}}
& $k=1$ & 1.24 & \textbf{2.11} & 1.38 & \textbf{2.19} & 1.74 & \textbf{1.83} & 1.62 & 1.48 & \textbf{1.99}\\
& & $k=3$ & \textbf{3.31} & 3.19 & 2.24 & 3.44 & \textbf{3.80} & \textbf{3.26} & 2.84 & 2.91 & \textbf{3.21}\\
& & $k=5$ & 3.72 & \textbf{4.29} &2.64& 3.88 & \textbf{5.19} & 3.78 & \textbf{4.05} & 3.06 & \textbf{4.22}\\
& & $k=10$ & 6.11 & \textbf{6.50} & 3.94 & 6.53 & \textbf{7.30} & \textbf{6.16} & 5.28 & 5.82 & \textbf{6.29}\\

\midrule

\multirow{4}{*}{\rotatebox[origin=c]{90}{FSOD}} &
\multirow{4}{*}{\rotatebox[origin=c]{90}{DeFRCN}}
& $k=1$  & \textbf{2.46} & 2.22 & 2.67 & 2.96 & \textbf{3.57} & \textbf{4.97} & 2.29 & 0.875 & \textbf{3.99}\\
& & $k=3$  & 4.97 & \textbf{5.26} & 6.13 & 5.41 & \textbf{6.81} & \textbf{7.93} & 7.60 & 2.15 & \textbf{7.60}\\
& & $k=5$  & \textbf{10.69} & 10.27 & 6.83 & \textbf{16.95} & 8.90 & \textbf{13.48} & 12.70 & 2.56 & \textbf{9.78}\\
& & $k=10$ & 12.82 & \textbf{13.62} & 12.23 & \textbf{18.18} & 11.48 & \textbf{17.45} & 15.96 & 5.37 & \textbf{12.49}\\

\midrule

\multirow{4}{*}{\rotatebox[origin=c]{90}{FSOD}} &
\multirow{4}{*}{\rotatebox[origin=c]{90}{YOLACT}}
& $k=1$ & 1.36 & \textbf{2.23} & 1.49 & \textbf{2.16} & 1.93 & \textbf{2.06} & 1.72 & 1.71 & \textbf{2.10}\\
& & $k=3$ & \textbf{3.42} & 3.36 & 2.51 & 3.42 & \textbf{4.05} & \textbf{3.44} & 3.05 & 3.30 & \textbf{3.33}\\
& & $k=5$ & 3.99 & \textbf{4.51} & 3.02 & 3.80 & \textbf{5.24} & 3.78 & \textbf{4.31} & 3.40 & \textbf{4.40}\\
& & $k=10$ & 6.19 & \textbf{6.70} & 4.43 & 6.36 & \textbf{7.31} & \textbf{6.18} & 5.60 & 6.09 & \textbf{6.45}\\

\bottomrule
\end{tabular}
\label{table:fine_grained_analysis}
\end{table*}

\noindent All fine-grained analysis results for YOLACT are shown in Table~\ref{table:fine_grained_analysis}. 

With respect to \emph{image quality} and \emph{object size}, our findings reinforce those of prior work.  Specifically, like prior work~\cite{dataset_VizWizCaption}, the algorithm typically performs better on images with higher quality.  Like other prior work~\cite{fsod_FSRW,OD_review}, algorithms typically perform worse for smaller objects.  However, our findings extend those reported in~\cite{fsod_FSRW,OD_review} since we define object sizes differently; i.e., they use smaller thresholds of $32^{2}$ and $96^{2}$. To our knowledge, our work is the first to offer insights into performance on larger objects due to the novel presence of such larger instances in our dataset. 

With respect to the \emph{presence of text}, overall the performance is slightly better for instances that contain text. Initially, we found this surprising.  We expected the opposite trend since we suspected that the limited prevalence of text in prior datasets would have led algorithm designers to not consider the presence of text in their algorithm designs. We suspect part of the reason for our finding is that, if the text on the objects is clear enough to be visible, then the image is high quality.  Additionally, the high frequency information from text regions in instance segmentations may be valuable predictive cues, despite the absence of the ability to recognize the text as text.  Finally, the presence of text has a strong correlation with particular categories, which may influence our findings. 

Finally, with respect to the \emph{presence of holes}, the performance is consistently worse for objects that contain holes.  The presence of holes raises the task complexity dramatically by requiring algorithms to go beyond locating object boundaries to also have a semantic understanding of all pixels within the object boundary.  According to our analysis in Section~\ref{sec:dataset} and the Supplementary Materials, objects with larger sizes tend to have more coverage by holes, including due to occlusion.  Therefore, we suspect that the poor performance that we observed for larger sized objects could be correlated with the poor performance we observing with our analysis here on objects with holes.

\subsection{Few-Shot Object Detection}
\label{subsec:algo_FSOD}

We benchmarked two FSOD algorithms for which code is publicly-available.  First, we chose Decoupled Faster R-CNN (\emph{DeFRCN}) with its default hyperparameters~\cite{fsod_DeFRCN}, since it is the state-of-the-art FSOD model.  It follows a two-stage fine-tuning paradigm. For our $k$-shot experiments, we randomly sample $k$ images to use for fine-tuning the model.   We also benchmark the YOLACT model used for FSIS by converting its segmentation results into bounding boxes. 

\vspace{-0.5em}\paragraph{Overall performance:}
Overall results are shown in Table~\ref{table:overall_performance}. These results resemble those observed for FSIS.  Specifically, both algorithms perform poorly on our dataset and much worse on our dataset than reported for the original dataset on which they were evaluated.  These results reinforce that our new dataset offers distinct challenges from existing datasets for FSOD algorithms.

\vspace{-0.5em}\paragraph{Fine-grained analysis:}
We perform the same fine-grained analysis conducted for FSIS with the two benchmarked FSOD models, and results are also reported in Table~\ref{table:fine_grained_analysis}.  While we observe that the level of \emph{image quality} does not correlate with algorithm performance, we do observe performance trends for the other three factors.  Moreover, these trends match those discussed for FSIS.  Specifically, both benchmarked models tend to perform the worst for small objects, perform better when text is present, and perform worse when holes are present.

\begin{table}[t!]
\begin{center}
\caption{Generalization of models trained on MS COCO for few-shot object detection to matching categories in our VizWiz-Fewshot-OD dataset.}
\label{table:cross_dataset}
\begin{tabular}{llrrrrrrrr}
\toprule
& & \multicolumn{4}{c}{mAP} & \multicolumn{4}{c}{mAP\textsubscript{50}} \\
model & testing set & 1 & 3 & 5 & 10 & 1 & 3 & 5 & 10 \\

\midrule

\multirow{2}{*}{DeFRCN~~}
& MS COCO~~ & 6.63  & 12.32 & 14.20  &16.69 & 12.50 & 21.69& 24.87 & 29.15 \\
& VizWiz & 1.32 & 3.43 & 2.17 & 4.57 & 2.74 & 5.86 & 3.39 & 6.53\\

\bottomrule
\end{tabular}
\end{center}
\end{table}

\vspace{-0.5em}\paragraph{Cross-dataset analysis:}
Finally, we evaluated DeFRCN's generalization performance across datasets.\footnote{Of note, we also conducted cross-dataset experiments with YOLACT in the FSIS and FSOD settings however the cross-dataset performance was negligible.  We attribute it to unsuccessful training with the chosen hyperparameters, both because the loss plateaued rather than converging with the new YOLACT hyperparameter values used in this paper and the loss exploded when using the original YOLACT values (i.e., the performance of YOLACT reported in the original paper could not be replicated when using the different set of training categories from MS COCO). In summary, the cross-dataset analysis results of YOLACT reinforce our initial findings that YOLACT performance is extremely sensitive to chosen hyperparameters and the training data, with custom tuning for each change.} To do, so we randomly selected $20$ of the $37$ categories found in both MS COCO and VizWiz-FewShot-OD-${25}^{i}$ as novel classes.  Next, we trained DeFRCN on the remaining $60$ MS COCO classes and then fine-tuned it with $k$-shot images randomly sampled from the $20$ novel classes in MS COCO.  The resulting model was evaluated on both the MS COCO test set as well as our VizWiz-FewShot-OD-${25}^{i}$ test set.  Results are shown in Table \ref{table:cross_dataset}. We observe significant gaps between scores on MS COCO and our dataset revealing that the algorithm generalizes poorly when encountering the domain shift between the two datasets. These findings reinforce that images in our dataset offers distinct algorithmic challenges from those observed in MS COCO.

\section{Conclusions}
\label{sec:conclusions}
We introduce the VizWiz-FewShot dataset to facilitate the community in designing few-shot learning models for object detection and instance segmentation that work well for the diverse set of challenges that emerge in real-world applications.  Our benchmarking of top few-shot localization algorithms reveal that valuable directions for future work are to better support objects that contain holes, very small and very large objects, and objects that lack text.

\paragraph{\bf Acknowledgments.}
This project was supported in part by a National Science Foundation SaTC award (\#2148080) and gift funding from Microsoft AI4A. We thank Leah Findlater and Yang Wang for contributing to this research idea and the anonymous reviewers for their valuable feedback to improve this work.

\newpage
\section*{\Large Appendix}
This document supports Sections 3 and 4 of the main paper. In particular, it includes the following:
\begin{itemize}
    \item List of categories in VizWiz-FewShot (supplements \emph{Section 3.1})
    \item Annotation interfaces (supplements \emph{Section 3.1})
    \item Quality control mechanisms for crowdsourcing instance segmentations (supplements \emph{Section 3.1})
    \item Comparison of unique categories with backward compatible categories  (supplements \emph{Section 3.2})
    \item Comparison of instance sizes in VizWiz-FewShot-IS-25\textsuperscript{i} with COCO-20\textsuperscript{i} (supplements \emph{Section 3.2})
    \item Fine-grained analysis of holes in instance segmentations (supplements \emph{Section 3.2})
    \item Analysis of the prevalence of instances and categories (supplements \emph{Section 3.2})
    \item Examples of our benchmarked algorithm's few-shot object detections on VizWiz-FewShot-OD-25\textsuperscript{i} (supplements \emph{Section 4.1})
\end{itemize}

\setcounter{section}{0}
\section{List of Categories in VizWiz-FewShot}

The $100$ categories in VizWiz-FewShot is divided into 4 folds in VizWiz-FewShot-OD-25\textsuperscript{i} and VizWiz-FewShot-IS-25\textsuperscript{i}. Our dataset includes object categories that overlaps with those in COCO-20\textsuperscript{i}~\cite{fsss_FeatureWeighting}, PASCAL-5\textsuperscript{i}~\cite{fsss_oneshot}, FSOD~\cite{fsod_attenRPN}, and FSS-1000~\cite{dataset_FSS1000} as well as categories that are unique to our dataset. In Table~\ref{table:all_categories}, we list all the categories in the 4 folds and indicate which categories are backward compatible with other few-shot datasets.

\begin{table}[t!]
\caption{List of the 100 categories in VizWiz-FewShot and the 4-fold class splits in VizWiz-FewShot-OD-25\textsuperscript{i} and VizWiz-FewShot-IS-25\textsuperscript{i}. The category that are backward compatible with other few-shot datasets are indicated as below. $\bullet$ refers to the overlapping categories with COCO-20\textsuperscript{i}~\cite{fsss_FeatureWeighting}, $\blacktriangle$ refers to the overlapping categories with PASCAL-5\textsuperscript{i}~\cite{fsss_oneshot}, $\blacksquare$ refers to the overlapping categories with FSOD~\cite{fsod_attenRPN}, and $\blacklozenge$ refers to the overlapping categories with FSS-1000~\cite{dataset_FSS1000}.}
\label{table:all_categories}
\begin{center}
\begin{tabular}{llll}
\hline\noalign{\smallskip}
$i=0~~~~~~~~~~~~~~~~~~$ & $i=1~~~~~~~~~~~~~~~~~~$ & $i=2~~~~~~~~~~~~~~~~~~$ & $i=3~~~~~~~~~~~~~~~~~~$\\
\noalign{\smallskip}
\hline
couch $\bullet$ & curtain $\blacksquare$ & sign & laptop $\blacklozenge$  $\bullet$ $\blacksquare$\\
watch $\blacksquare$ & spoon $\bullet$ $\blacksquare$ $\blacklozenge$ & truck $\bullet$ $\blacksquare$ & cell phone \\
drawer $\blacksquare$ & food menu & vacuum $\blacklozenge$ & bed $\bullet$ \\
landline phone & bowl $\bullet$ $\blacksquare$ & clock $\bullet$ & vase $\bullet$ $\blacksquare$ $\blacklozenge$ \\
strawberry $\blacksquare$ $\blacklozenge$ & receipt & ipad $\blacklozenge$ & dial \\
painting & newspaper & perfume $\blacksquare$ $\blacklozenge$ & broccoli $\bullet $ $\blacksquare$ $\blacklozenge$ \\
pillow $\blacksquare$ $\blacklozenge$ & album & towel $\blacksquare$ & scale $\blacksquare$ \\
crockpot & cereal box & stool $\blacksquare$ $\blacklozenge$ & piano \\
pen $\blacksquare$ $\blacklozenge$ & lamp $\blacksquare$ & microwave $\blacklozenge$  $\bullet $ $\blacksquare$ & oven  $\bullet $ $\blacksquare$ \\
speaker $\blacklozenge$ & dog  $\bullet $ $\blacktriangle$ & magazine & house \\
ring & sandwich  $\bullet $ $\blacklozenge$ & gift card & bottle  $\bullet $ $\blacktriangle$ $\blacksquare$ \\
sock $\blacksquare$ $\blacklozenge$ & ceiling fan $\blacklozenge$ & stapler $\blacksquare$ $\blacklozenge$ & hat \\
car  $\bullet $ $\blacktriangle$ & toaster  $\bullet $ $\blacksquare$ $\blacklozenge$ & dog collar & bird  $\bullet $ $\blacktriangle$\\
fork  $\bullet $ $\blacksquare$ $\blacklozenge$ & cake  $\bullet $ $\blacksquare$ & sandal $\blacksquare$ $\blacklozenge$ & knife  $\bullet $ $\blacksquare$ $\blacklozenge$ \\
coin $\blacksquare$ $\blacklozenge$ & book  $\bullet $ $\blacksquare$& shoe & wallet $\blacksquare$ $\blacklozenge$ \\
envelope $\blacksquare$ $\blacklozenge$ & apple  $\bullet $ $\blacksquare$ $\blacklozenge$ & banana  $\bullet $ $\blacksquare$ $\blacklozenge$ & suitcase  $\bullet $ $\blacksquare$ $\blacklozenge$ \\
bracelet $\blacklozenge$ & packet $\blacksquare$ & carrot  $\bullet $ $\blacksquare$ $\blacklozenge$ & backpack  $\bullet $ $\blacksquare$ $\blacklozenge$ \\
television $\blacksquare$ $\blacklozenge$ & rug & microphone $\blacklozenge$ & sweatshirt $\blacklozenge$ \\
cash & printer $\blacksquare$ $\blacklozenge$ & plate $\blacksquare$ $\blacklozenge$ & stove $\blacklozenge$ \\
monitor $\blacktriangle$ $\blacklozenge$ & electric fan $\blacklozenge$ & sticker & computer mouse   $\bullet $ $\blacksquare$ $\blacklozenge$ \\
sink  $\bullet $ $\blacksquare$ & remote  $\bullet $ & person  $\bullet $ $\blacktriangle$ & wine $\blacksquare$ \\
guitar $\blacksquare$ $\blacklozenge$ & orange  $\bullet $ $\blacksquare$ $\blacklozenge$ & chair  $\bullet $ $\blacktriangle$ & laundry machine \\
purse & ramen & refrigerator  $\bullet $ $\blacksquare$ $\blacklozenge$ & keyboard   $\bullet $ $\blacklozenge$ \\
calculator $\blacksquare$ $\blacklozenge$ & cup  $\bullet $ $\blacklozenge$ & bar & key $\blacklozenge$ \\
cat  $\bullet $ $\blacktriangle$ $\blacksquare$ & flashdrive & pizza  $\bullet $ $\blacksquare$ $\blacklozenge$ & tube \\
\noalign{\smallskip}
\hline
\end{tabular}
\end{center}
\end{table}

\section{Annotation Interfaces}
We utilized two interfaces for collecting annotations from Amazon Mechanical Turk crowdworkers. 

\begin{figure}[t!]
    \centering
    \includegraphics[width=1.0\textwidth]{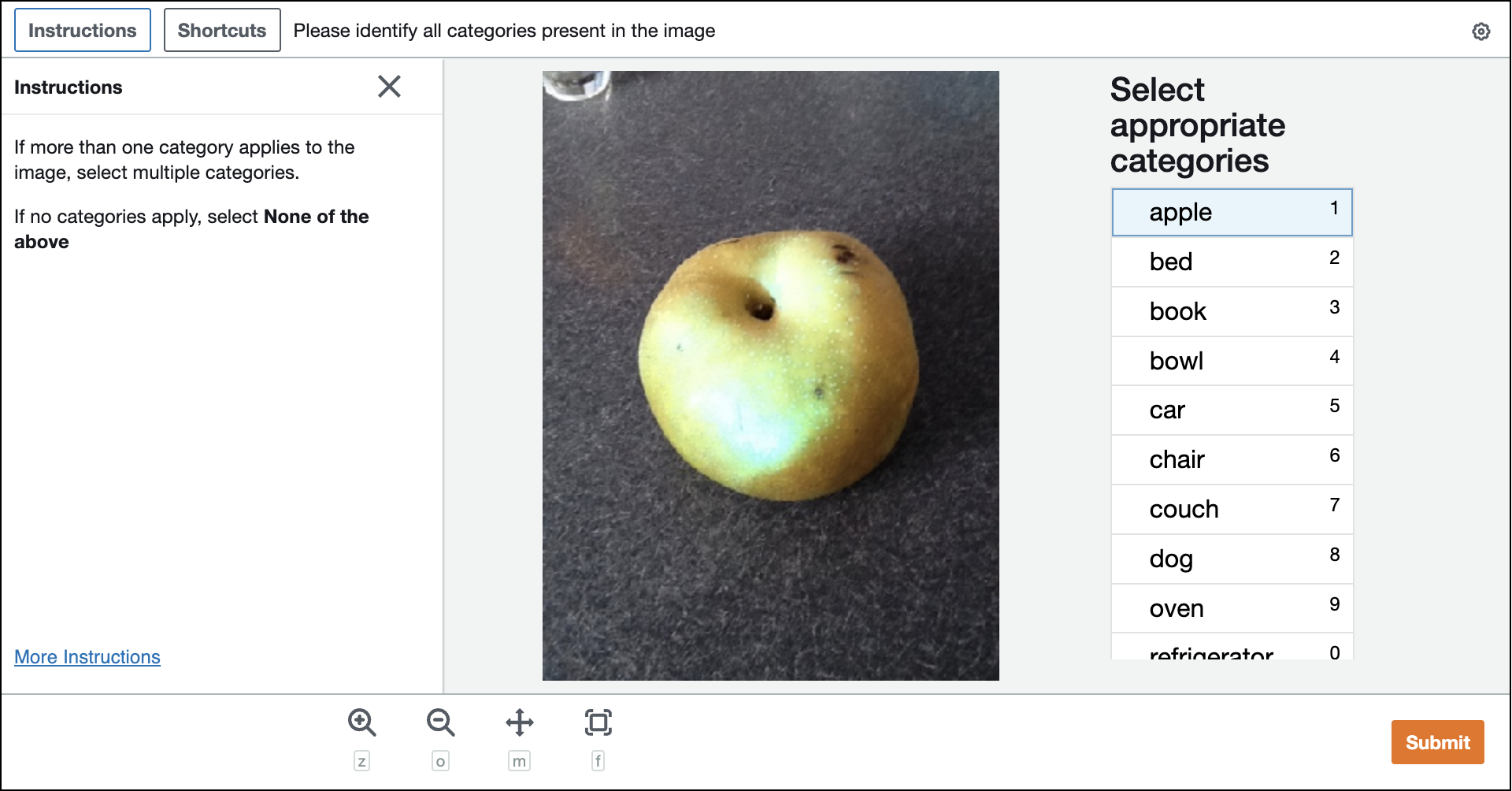}
    \caption{Our image classification interface. We provide instructions, an image, and a multi-select list that shows $\sim25$ categories on the right.}
    \label{fig:image_classification_interface}
\end{figure}

The first interface is for image classification, and a screenshot of it is shown in Figure~\ref{fig:image_classification_interface}. It shows instructions on the left side indicating to select all categories present in the image or \textit{None of the above}.  The image is displayed in the center of the interface and approximately 25 categories to select from are displayed on the right side.

\begin{figure}[t!]
    \centering
    \includegraphics[width=1.0\textwidth]{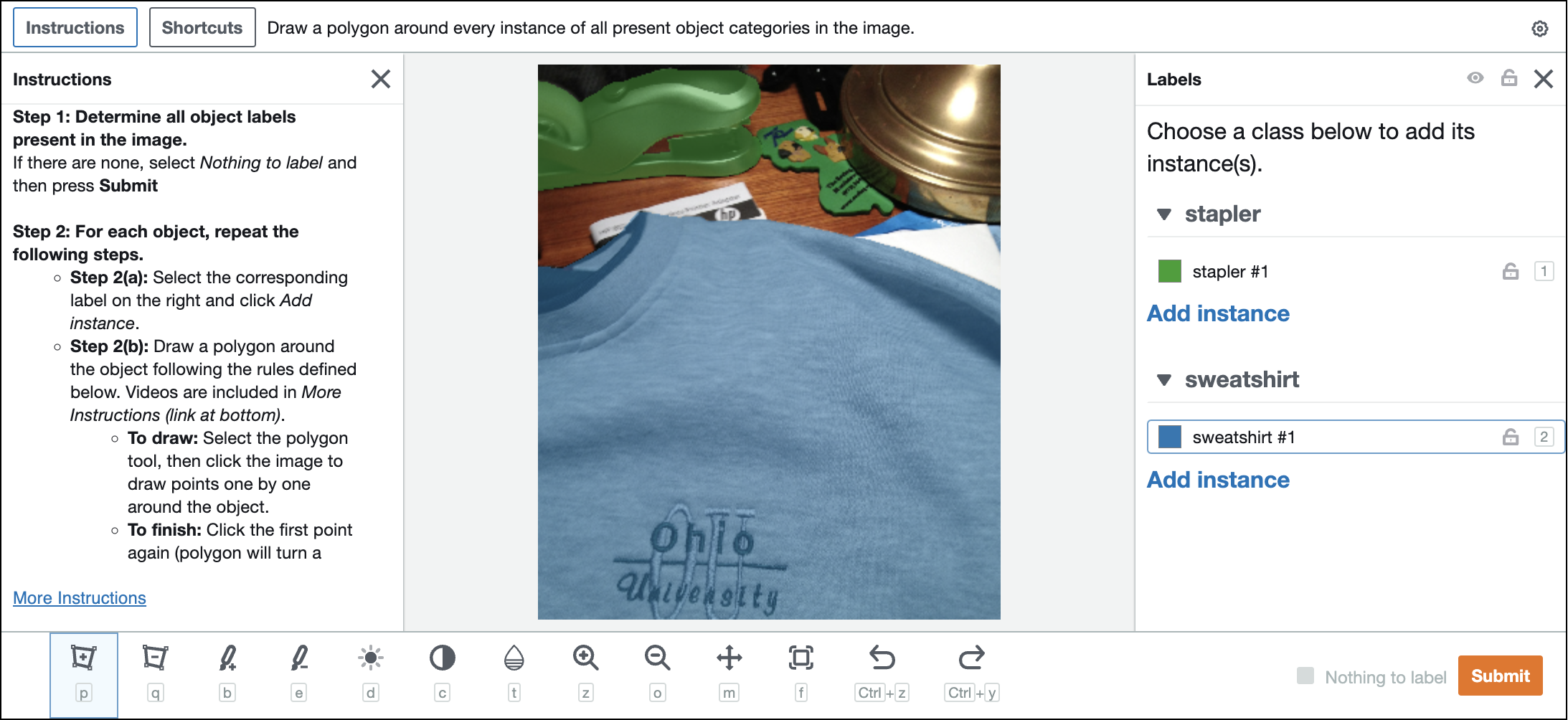}
    \caption{Our instance segmentation interface. We provide instructions, an image, and a few categories for which annotators should locate all instances.}
    \label{fig:instance_segmentation_interface}
\end{figure}

\begin{figure}[t!]
    \centering
    \includegraphics[width=0.54\textwidth]{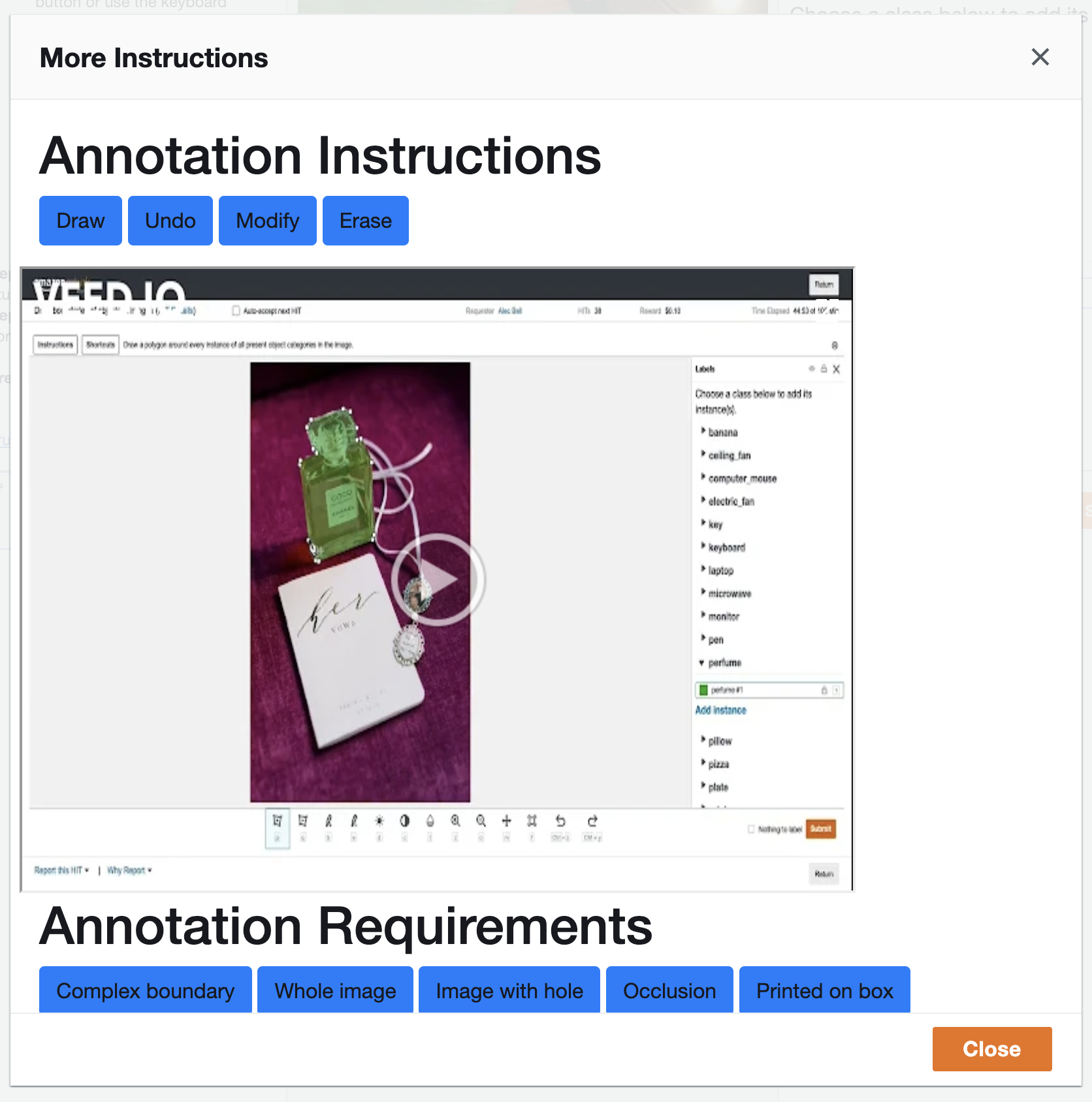}
    \includegraphics[width=0.41\textwidth]{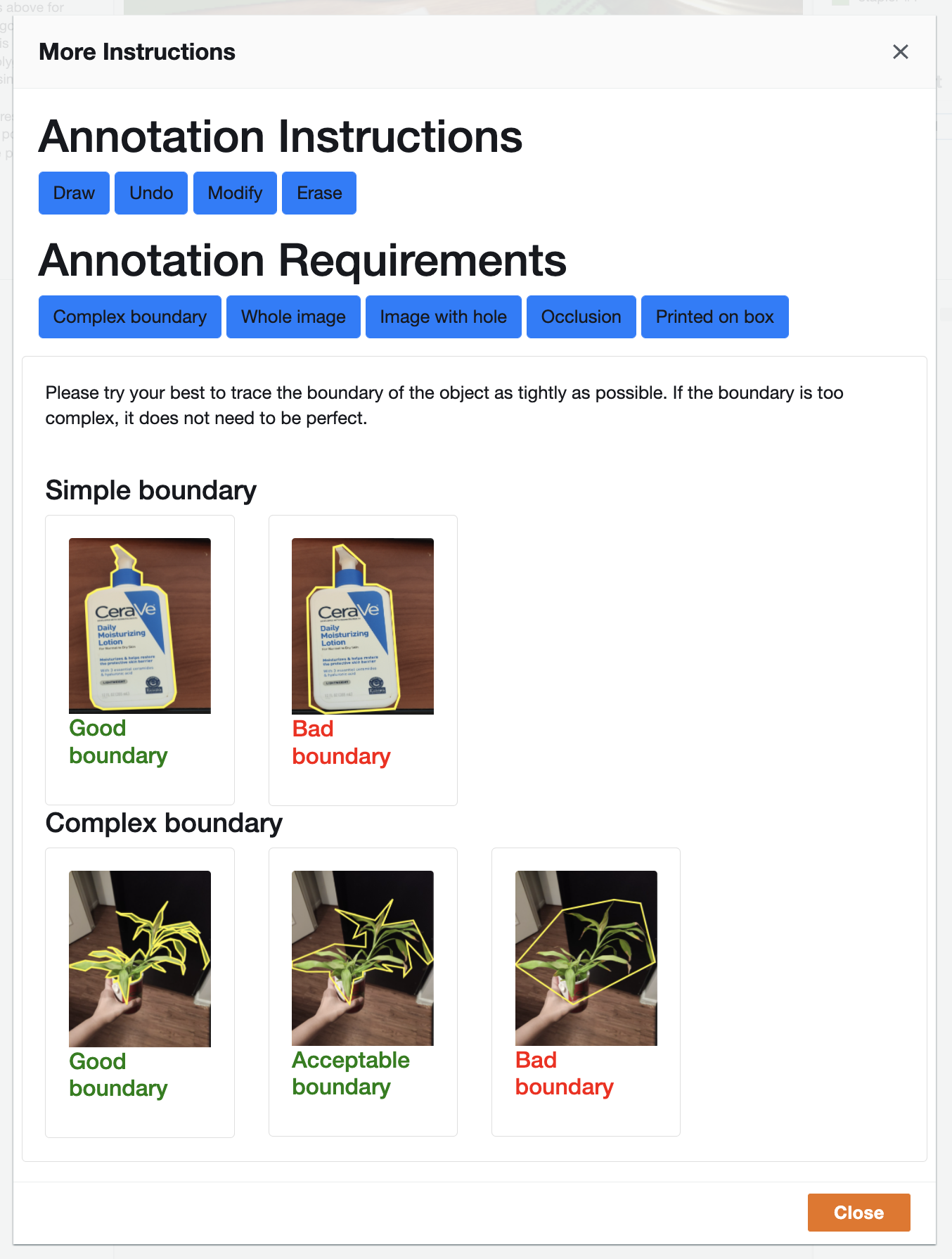}
    \caption{Additional instructions for our instance segmentation task showing short video tutorials for how to use the annotation tools (on the left) and indicating how to handle edge-case scenarios (on the right).}
    \label{fig:instance_segmentation_annotation_instr_req}
\end{figure}

The second interface is for instance segmentation, and a screenshot of it is shown in Figure~\ref{fig:instance_segmentation_interface}. A step-by-step list of instructions is shown on the left of the interface and is listed in Figure~\ref{fig:instance_segmentation_instructions}.  It is supplemented with \textit{More Instructions}, which we partially show in Figure~\ref{fig:instance_segmentation_annotation_instr_req}.  The additional guidance includes links to videos explaining how to perform various annotation operations such as drawing a polygon, undoing an action, modifying an existing instance, and erasing an existing instance or polygon partially or entirely. The extra instructions also indicate how to handle edge-case scenarios, alongside examples of what to do and what not to do. Covered edge-case scenarios are how to handle complex boundaries, high image coverage, holes, and occlusions. We also specify that objects printed on boxes (such as pizza on a frozen pizza box) should be annotated. 

\begin{figure}[t!]
    \noindent\rule{\textwidth}{0.4pt}
    \begin{description}
    \vspace{-0.2em}
        \item[Step 1:] Determine all object labels present in the image. If there are none, select Nothing to label and then press Submit
        \item[Step 2:] For each object, repeat the following steps.
        \item[Step 2(a):] Select the corresponding label on the right and click Add instance.
        \item[Step 2(b):] Draw a polygon around the object following the rules defined below. Videos are included in More Instructions (link at bottom).
            \begin{description}
                \item[To draw:] Select the polygon tool, then click the image to draw points one by one around the object.
                \item[To finish:] Click the first point again (polygon will turn a color when it is fully connected), and then press the return key. If you need more segments for the same object, you can draw them from here and continue to press return after each one.
                \item[To undo:] Click the undo button or use the keyboard shortcut Ctrl+Z.
                \item[To erase:] Select the polygon eraser tool, then draw the area you wish to erase and press enter.
                \item[To fix a polygon:] Follow the draw/erase steps above for any existing polygon to make changes to it. This can be done after the polygon is finalized by pressing the return key.
            \end{description}
        \item[Step 2(c):] YOU MUST press the return key to finalize the polygon.
        \item[Step 2(d):] Verify that the polygon is stored for the instance.
        \item[Step 3:] Press Submit.
        \vspace{-0.2em}
    \end{description}
    \noindent\rule{\textwidth}{0.4pt}
    \caption{Instructions provided for our instance segmentation task.}
    \label{fig:instance_segmentation_instructions}
\end{figure}

\begin{figure}[t!]
\centering
\begin{tabular}{cccc}
\subcaptionbox{ceiling fan}{\includegraphics[width = 0.3\textwidth]{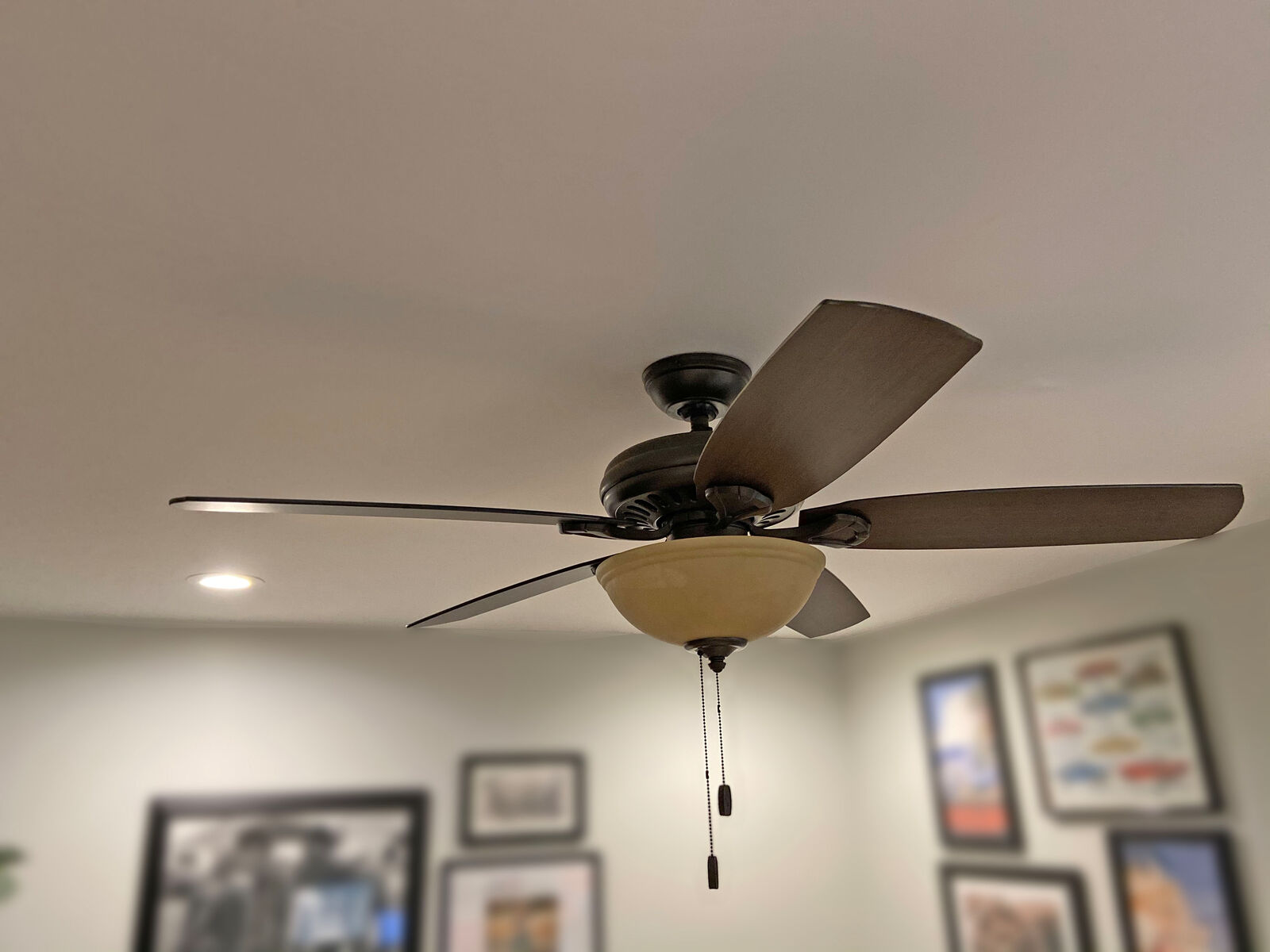}} &
\subcaptionbox{keyboard}{\includegraphics[width = 0.3\textwidth]{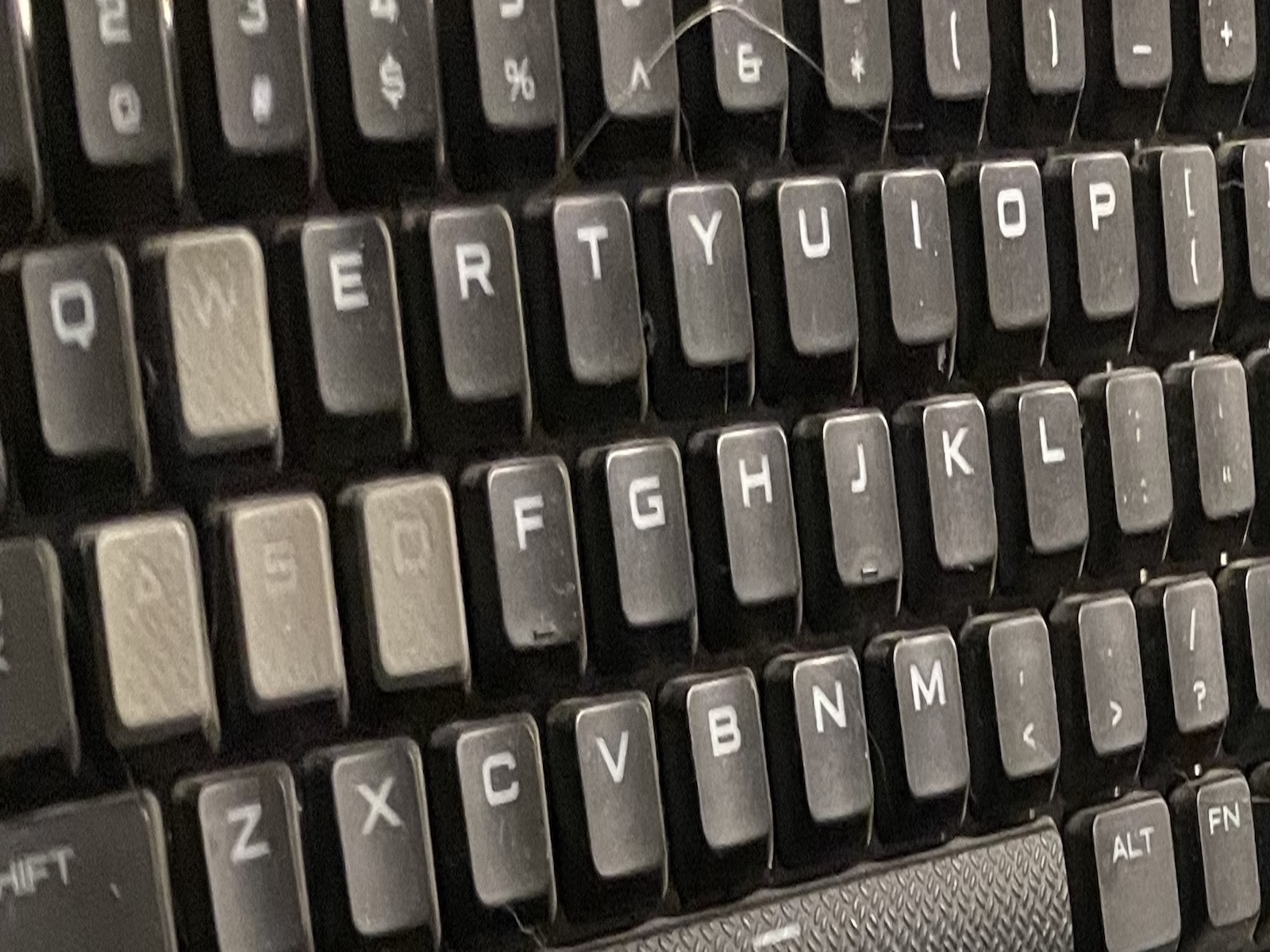}} &
\subcaptionbox{stove}{\includegraphics[width = 0.3\textwidth]{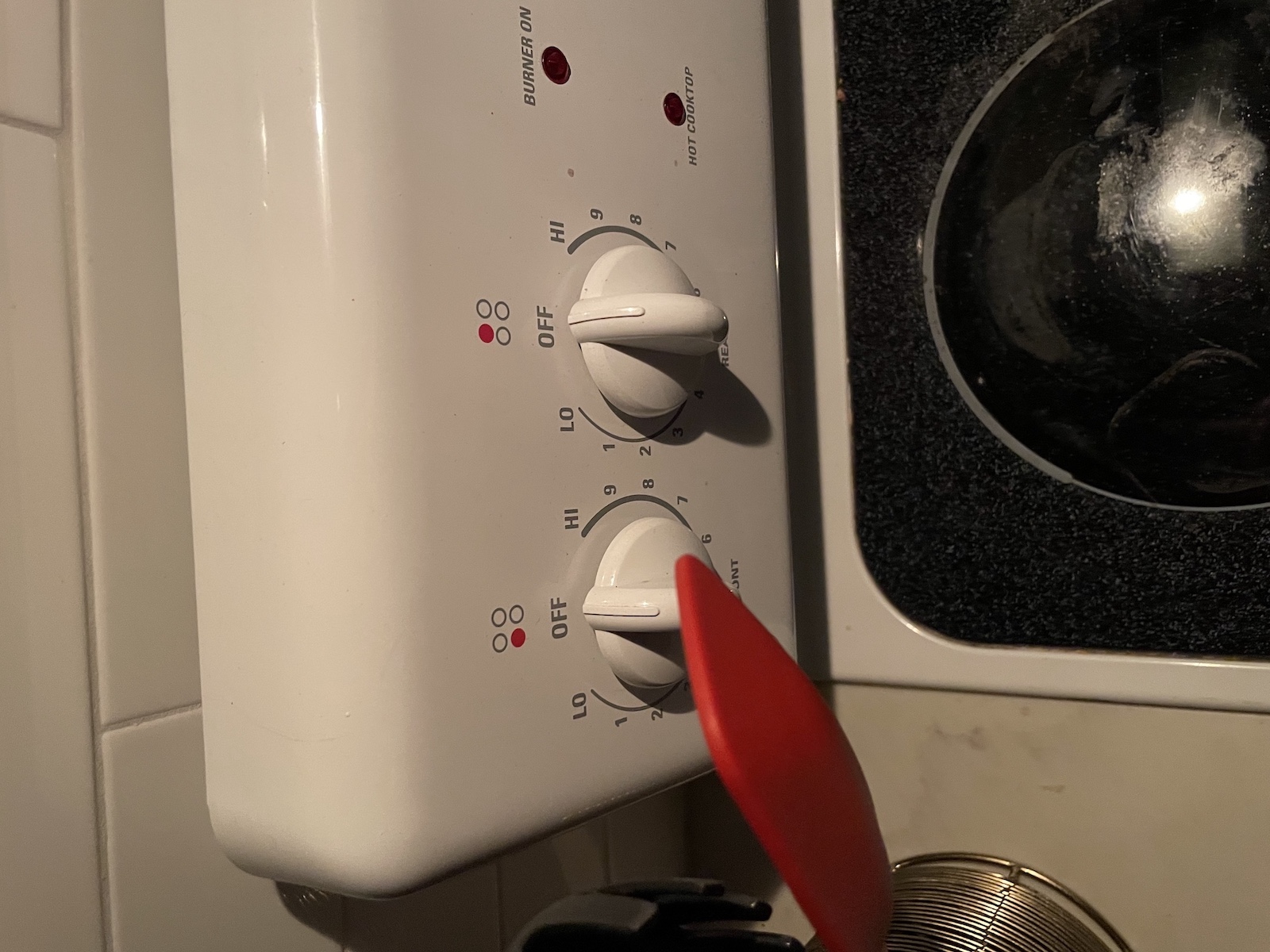}} \\
\subcaptionbox{keys}{\includegraphics[width = 0.3\textwidth]{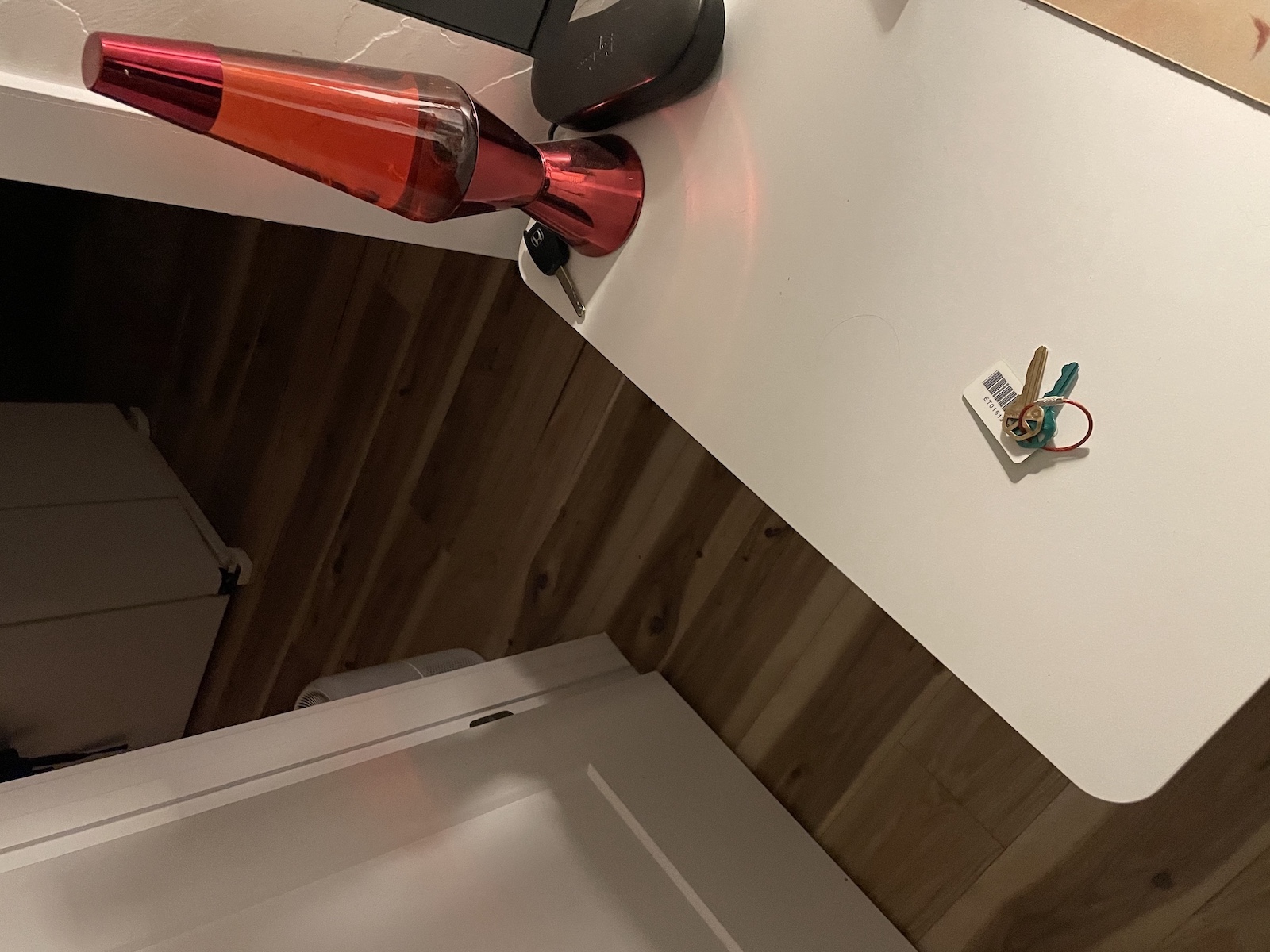}} &
\subcaptionbox{electric fan}{\includegraphics[width = 0.3\textwidth]{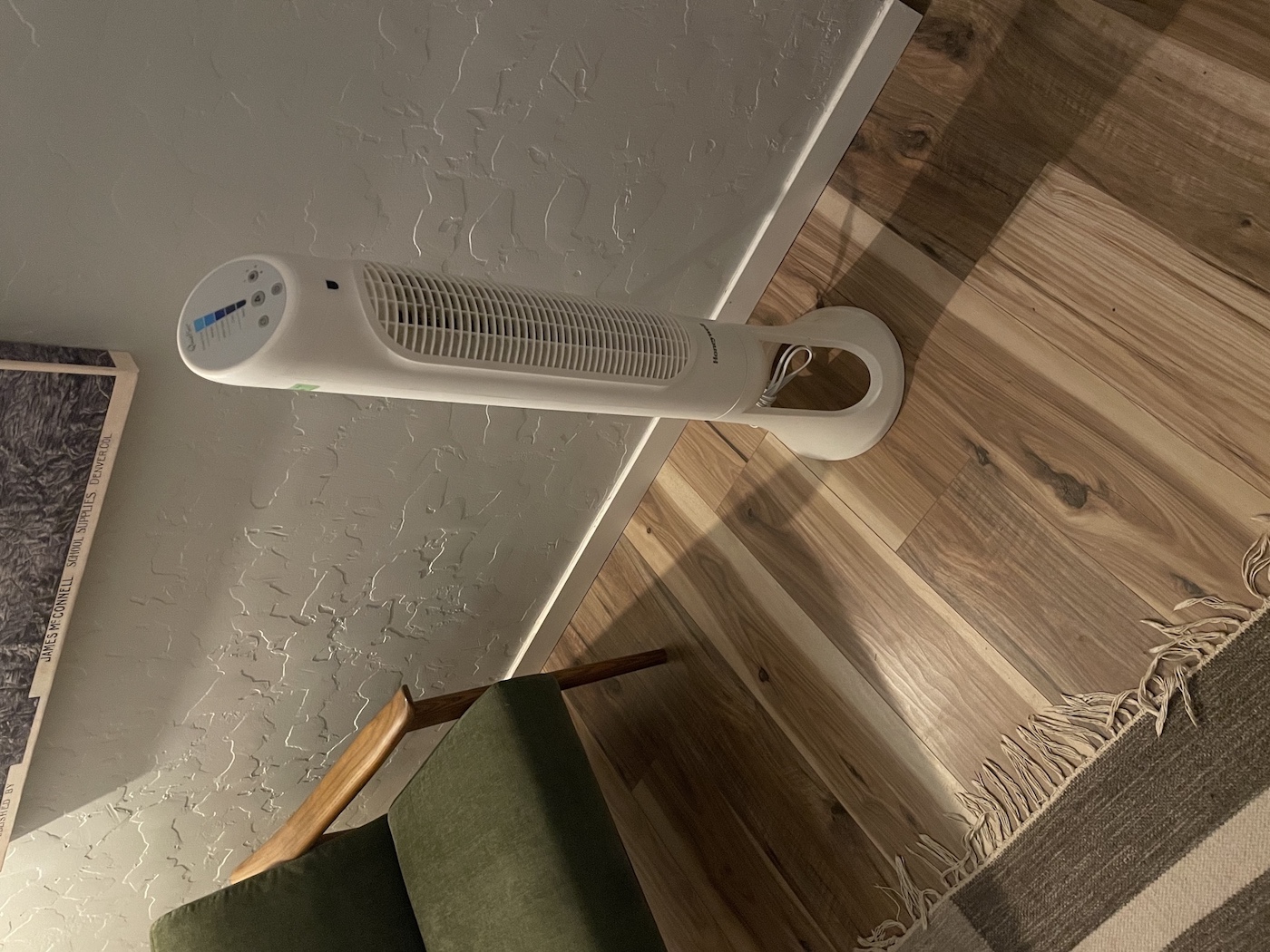}} &
\subcaptionbox{laptop and monitor}{\includegraphics[width = 0.3\textwidth]{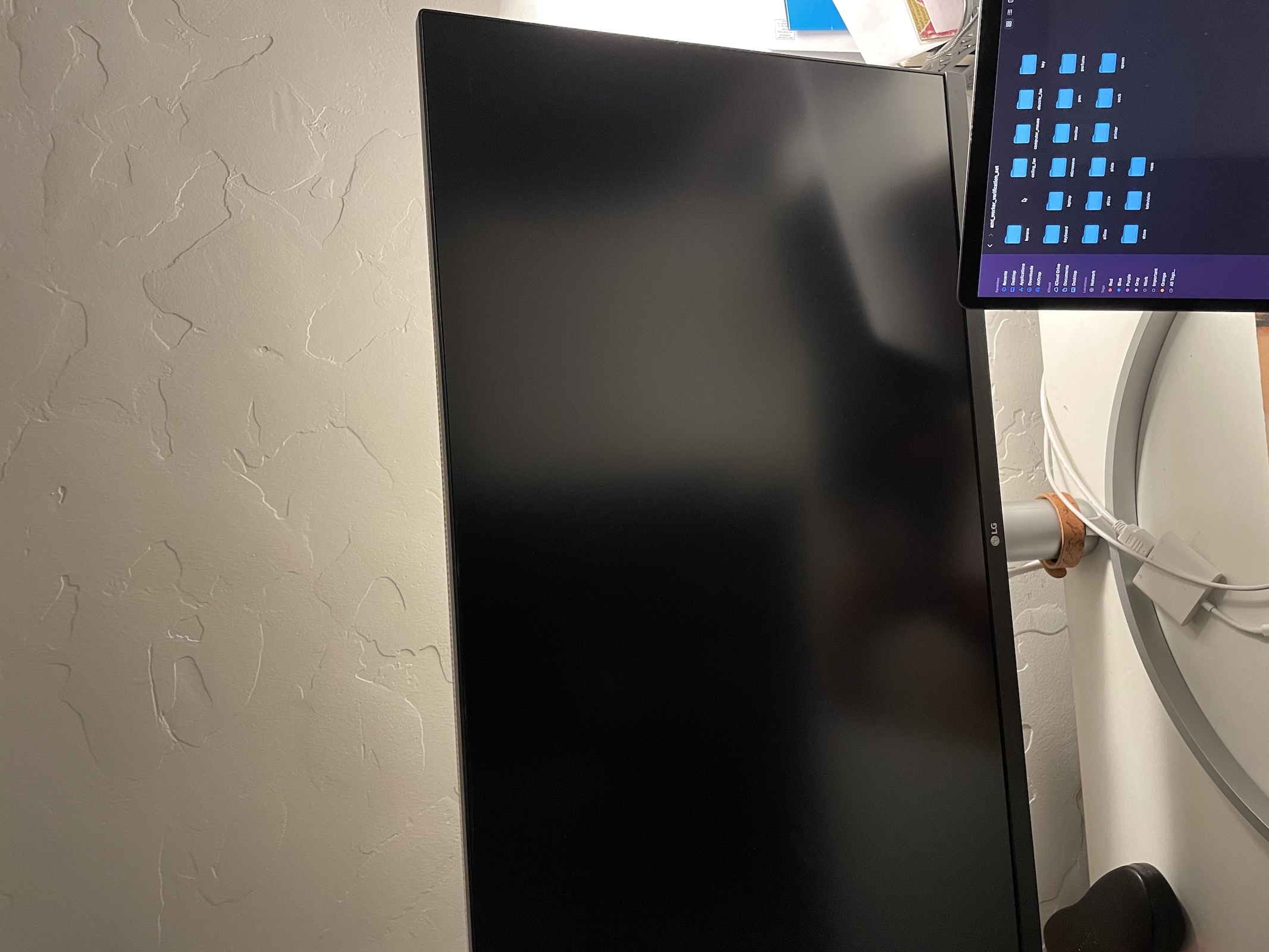}} \\
\subcaptionbox{spoon}{\includegraphics[width = 0.3\textwidth]{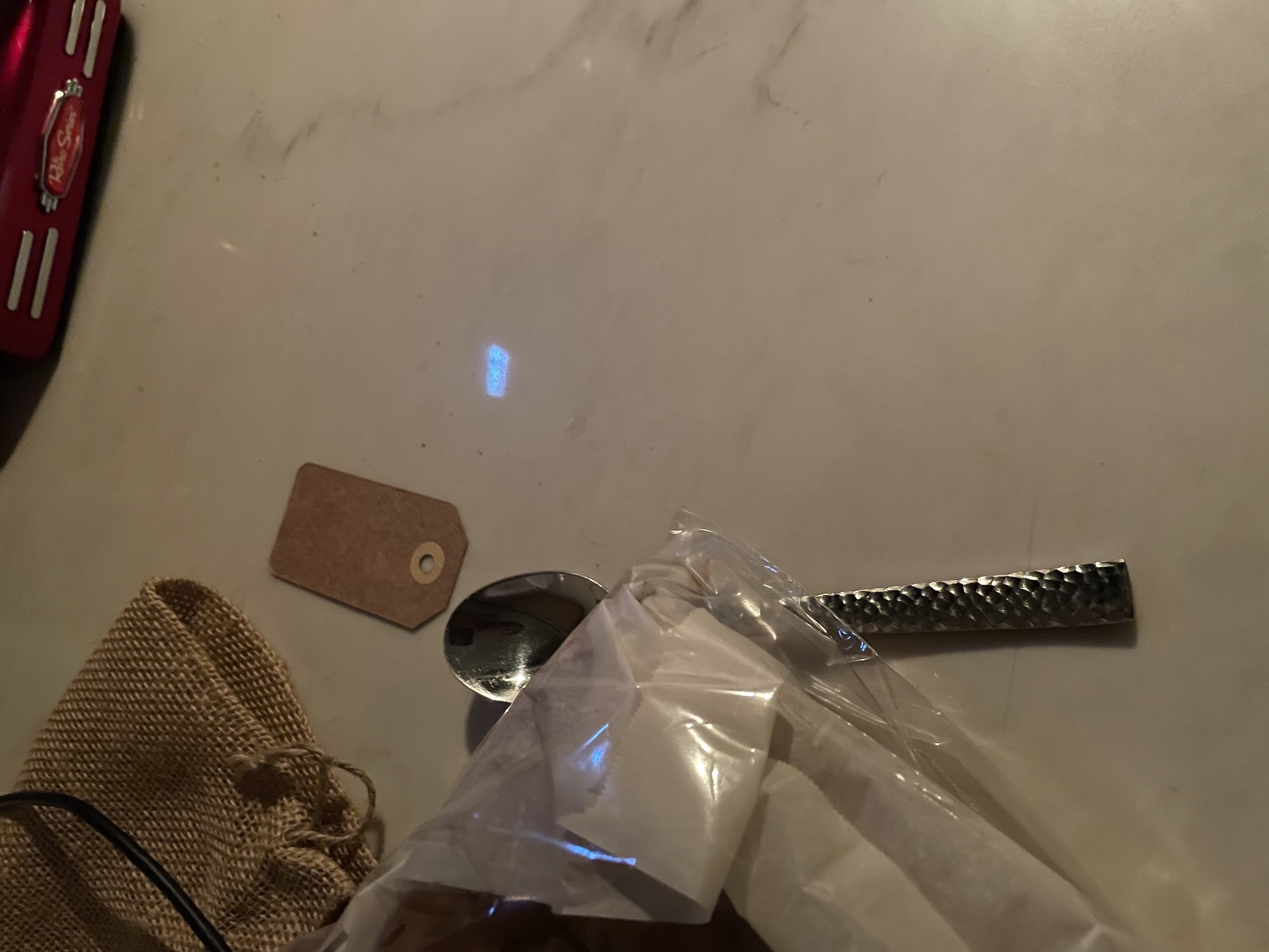}} &
\subcaptionbox{pizza}{\includegraphics[width = 0.3\textwidth]{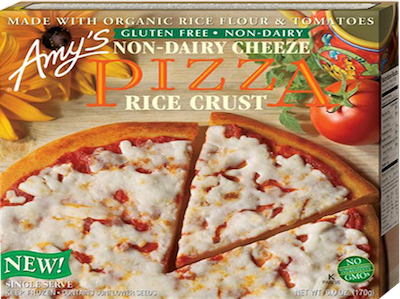}} &
\subcaptionbox{pizza}{\includegraphics[width = 0.3\textwidth]{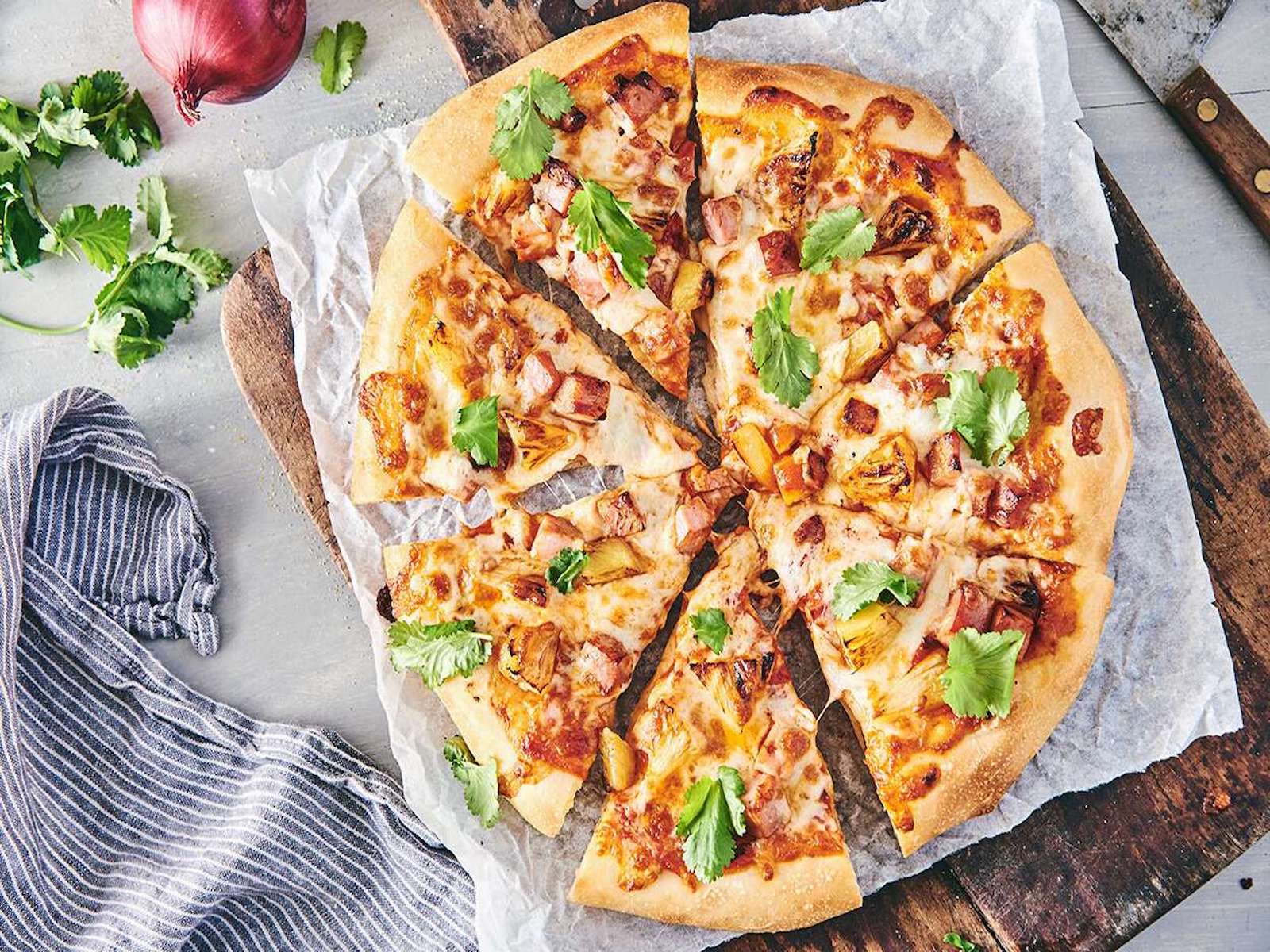}} &
\end{tabular}
\caption{The nine images that had to be correctly annotated by crowdworkers in their qualification test for our instance segmentation task.}
\label{fig:challenging}
\end{figure}

\section{Quality Control Mechanisms for Crowdsourcing}
In the main paper, we noted that crowdworkers had to pass a qualification test in order to work on our instance segmentation tasks.  The nine tasks that we presented in this qualification test are shown in Figure~\ref{fig:challenging}.  These photos were taken by the authors or found online.  The tasks test workers ability to annotate complex boundaries (images \textit{a,d}), the whole image (image \textit{b}), occlusions (images \textit{c,g}), holes in instances (image \textit{e}), and objects found both its original physical form and in print (images \textit{h,i}). 

\section{Comparison of Unique Categories in VizWiz-FewShot-IS-25 with COCO-20\textsuperscript{i} Categories}
We conduct fine-grained analysis of our dataset to elucidate whether categories that are unique to our dataset manifest different characteristics than categories that are in common with COCO-20\textsuperscript{i}~\cite{fsss_FeatureWeighting}. To do so, we analyze the boundary complexity, image coverage, and presence of text separately for the unique categories in our dataset and the 37 categories overlapping with those of COCO-20\textsuperscript{i}. Across all instance segmentations in each subset, we compute the mean and standard deviation of the isoperimetric inequality and the image coverage as well as the percentage of instance segmentations containing text.  

Results are shown in Table~\ref{table:text_prevalence}.  The key difference between the two subsets is that the presence of text on an instance is more prevalent for categories unique to our dataset.  We also observe for categories unique to our dataset that, on average, the objects' boundaries are slightly less complex (i.e., larger scores) and occupy slightly more of the images.

\begin{table}[t!]
\begin{center}
\begin{tabular}{lcc}
\hline\noalign{\smallskip}
 & Shared & Unique \\
\noalign{\smallskip}
\hline
\noalign{\smallskip}
Boundary Complexity & $0.51\pm0.21$ & $0.54\pm0.21$ \\
Image Coverage & $0.24\pm0.27$ & $0.23\pm0.27$ \\
Presence of Text & 17.69\% & 23.69\% \\
% \textbf{COCO-20\textsuperscript{i}} & 4.6\% & 3.45\% & 7.09\% \\
\hline
\end{tabular}
\end{center}
\caption{Comparison of instance segmentations in our dataset between those that show unique categories versus backward compatible categories in COCO-20\textsuperscript{i}.}
\label{table:text_prevalence}
\end{table}

\begin{table}[t!]
\begin{center}
\begin{tabular}{lcccccccc}
\hline\noalign{\smallskip}
Dataset & \multicolumn{3}{c}{Instance Sizes} \\
& small & medium & large \\
\noalign{\smallskip}
\hline
\noalign{\smallskip}
\textbf{Ours} & 2.11\% & 10.07\% & 87.82\% \\
\textbf{COCO-20\textsuperscript{i}} & 41.78\% & 34.92\% & 23.31\% \\
\hline
\end{tabular}
\end{center}
\caption{Proportion of instances belonging to each sizing category as defined by MS-COCO. We notice that the vast majority of instances in our dataset are large, while instance sizes are more evenly distributed in COCO.}
\label{table:sizes}
\end{table}

\section{Comparison of Instance Sizes in VizWiz-FewShot-IS-25\textsuperscript{i} with COCO-20\textsuperscript{i}}
We analyze what proportion of instance segmentations in our dataset fall into small, medium, and large sizes based on the thresholds proposed in MS COCO~\cite{dataset_coco}: $32^{2}$ and $96^{2}$.  Results are shown in Table~\ref{table:sizes}.  While these thresholds are able to roughly divide COCO-20\textsuperscript{i} into three even subsets, our dataset appears to have an extreme distribution where most of the instances fall into the large category. 

\section{Fine-grained Analysis of Holes in Instance Segmentations}

\paragraph{Limitation of prior work in lacking hole annotations:}

\begin{figure}[t!]
\centering
\includegraphics[width=1.0\textwidth]{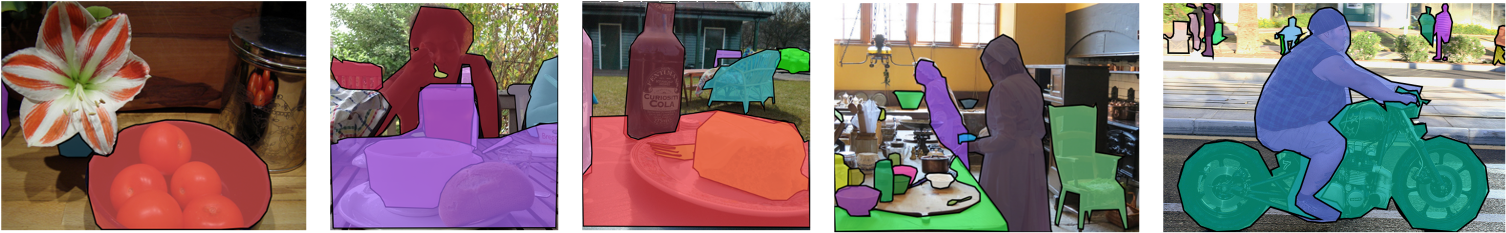}
\caption{Examples from COCO-20\textsuperscript{i} dataset shows how occluding objects are sometimes removed from instance segmentations (e.g., flower in first example) but sometimes not (e.g., oranges in the bowl in the first example).}
\label{img:no_holes_inconsistency}
\end{figure}

As noted in the main paper, COCO-$20^{i}$, instance segmentations lack holes whereas our dataset includes hole annotations. We note that the absence of holes leads to a limitation in prior work's approach for locating instance segmentations.  Specifically, content from occlusions is excluded from instance segmentations when the occlusions overlap their outer boundaries but included when the occlusions are fully enclosed in the instance segmentations.  In other words, there is an inconsistency in whether the instance segmentations include pixels that do not belong to the objects. This is exemplified in Figure~\ref{img:no_holes_inconsistency}, where the food contained in the bowl is included in the instance segmentation while the flower occluding the bowl is excluded from the instance segmentation.  Our dataset, in contrast, leads to a consistent definition of instance segmentations by annotating holes and so always excluding any pixels that do not belong to the target category. 

\begin{figure}[t!]
\center
\begin{subfigure}[t]{0.49\textwidth}
\centering
\includegraphics[width=1.0\textwidth]{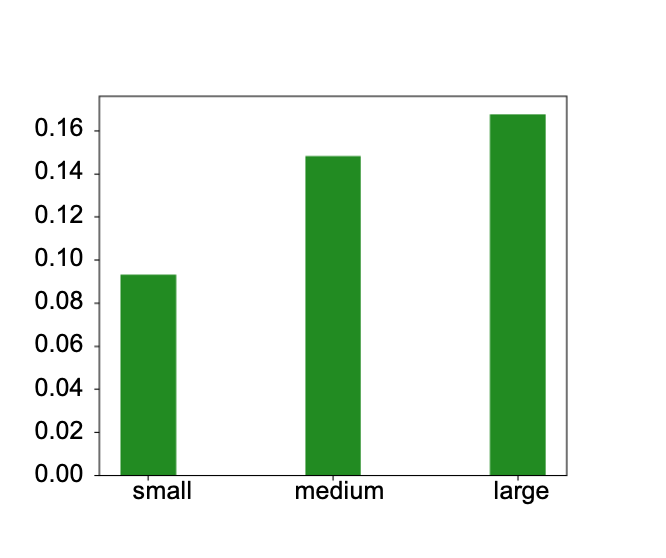}
\caption{Hole prevalence for objects of different sizes.}
\label{fig:holes_analysis_by_size_a}
\end{subfigure}
\begin{subfigure}[t]{0.49\textwidth}
\includegraphics[width=1.0\textwidth]{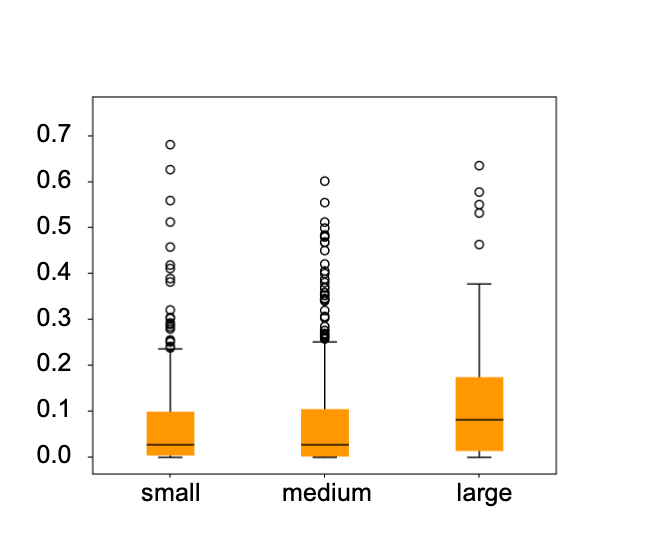}
\caption{Distribution of hole coverage for objects of different sizes.}
\label{fig:holes_analysis_by_size_b}
\end{subfigure}
\caption{Analysis on hole prevalence and hole coverage distribution in VizWiz-FewShot-IS\textsuperscript{i} by category. Only the instances with holes are included in the hole coverage distribution.}
\end{figure}

\begin{figure}[b!]
\center
\begin{subfigure}[t!]{1.0\textwidth}
\centering
\includegraphics[width=1.0\textwidth]{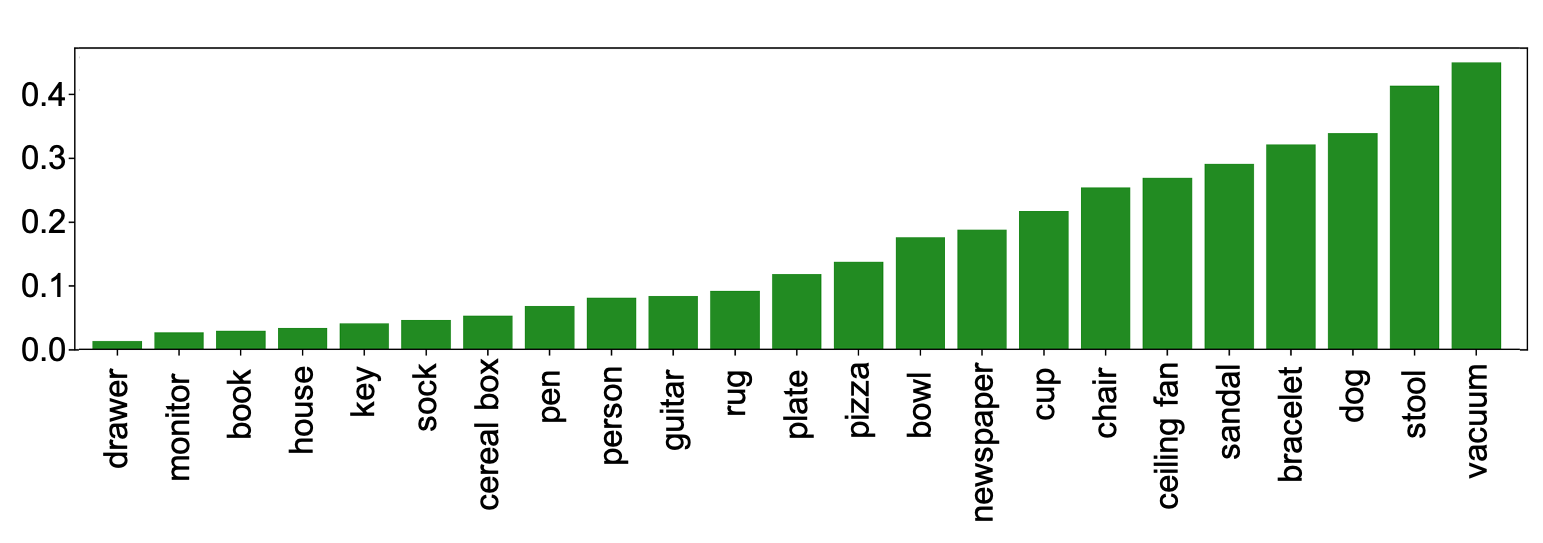}
\caption{Hole prevalence by category.}
\label{fig:holes_analysis_by_category_a}
\end{subfigure}
\begin{subfigure}[t!]{1.0\textwidth}
\includegraphics[width=1.0\textwidth]{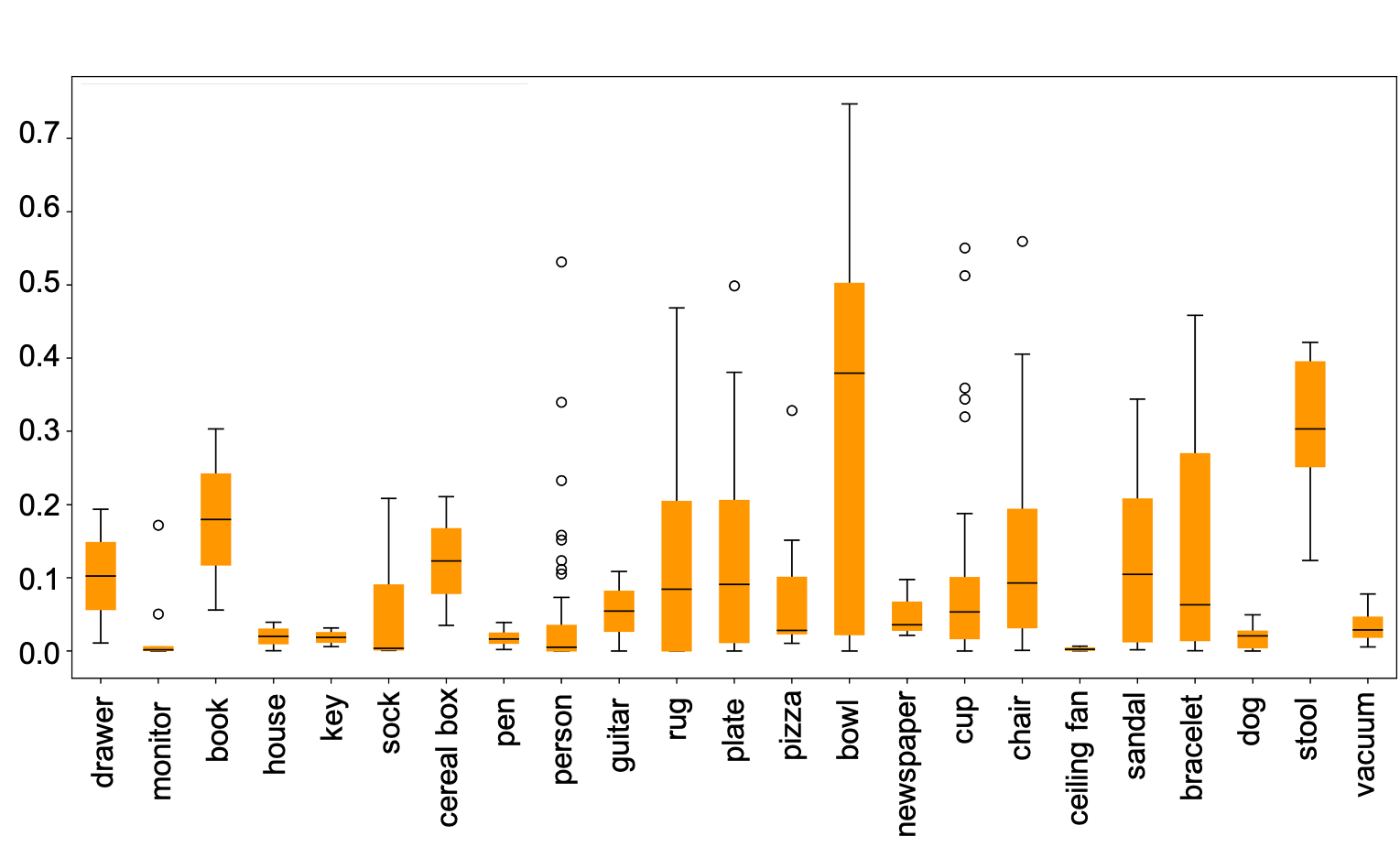}
\caption{The distribution of hole coverage by category.}
\label{fig:holes_analysis_by_category_b}
\end{subfigure}
\caption{Analysis of hole prevalence and hole coverage distribution in VizWiz-FewShot-IS\textsuperscript{i} by category. Only the instances with holes are included in the hole coverage distribution. The categories are sorted by hole prevalence in ascending order.}
\end{figure}

\paragraph{Hole prevalence:}
We conducted fine-grained analysis on the proportion of instance segmentations with holes based on object size.  To do so, we divided all instance segmentations into small, medium, and large buckets using as thresholds $200^2$ and $550^2$.  The holes with areas smaller than 10 pixels are excluded from consideration since we found that such annotations typically reflected holes that were accidentally/mistakenly created by the annotators. Some of these errors are due to the conversion of annotations from vertex-base to pixel-base. Results are shown in Figure~\ref{fig:holes_analysis_by_size_a}.  The trend shows that the proportion of instances with holes grows as object size grows.  We also report the proportion of instance segmentations with holes based on object category for a random sample of $23$ categories.  The results are shown in Figure~\ref{fig:holes_analysis_by_category_a}, sorted by the proportion of instances with holes in ascending order.  Altogether, these results highlight that the prevalence of holes differs on per-size and per-category bases.

\paragraph{Typical hole coverage:}
We also analyzed the percentage of pixels all holes occupy in each instance segmentation from all pixels contained in each instance segmentation (when including the hole pixels). The results of hole coverage divided by size and category are presented in Figure~\ref{fig:holes_analysis_by_size_b} and Figure~\ref{fig:holes_analysis_by_category_b} respectively. We observe that the proportion of hole coverage tends to be larger for larger objects. 

Comparing the hole prevalence and hole coverage results in Figure~\ref{fig:holes_analysis_by_category_a} and Figure~\ref{fig:holes_analysis_by_category_b}, we find that the proportion of instance segmentations with holes and hole coverage do not have a strong correlation in per-category bases. For example, a high percentage of stools and vacuums stools have holes ($41.4\%$ and $45.0\%$, respectively), but based on the natural structure of these categories, they show distinct distributions in hole coverage. On the other hand, the categories that tend to be occluded by other objects, such as bowls, might not as frequently include holes. However, this does not influence the high hole coverage of bowls, where foods are often found to cover a large area in our dataset. Fig.~\ref{fig:holes_examples} shows examples of stools, vacuums, and bowls.

\section{Analysis of the prevalence of instances and categories}

\begin{figure}[b!]
  \center
    \includegraphics[width=1.0\textwidth]{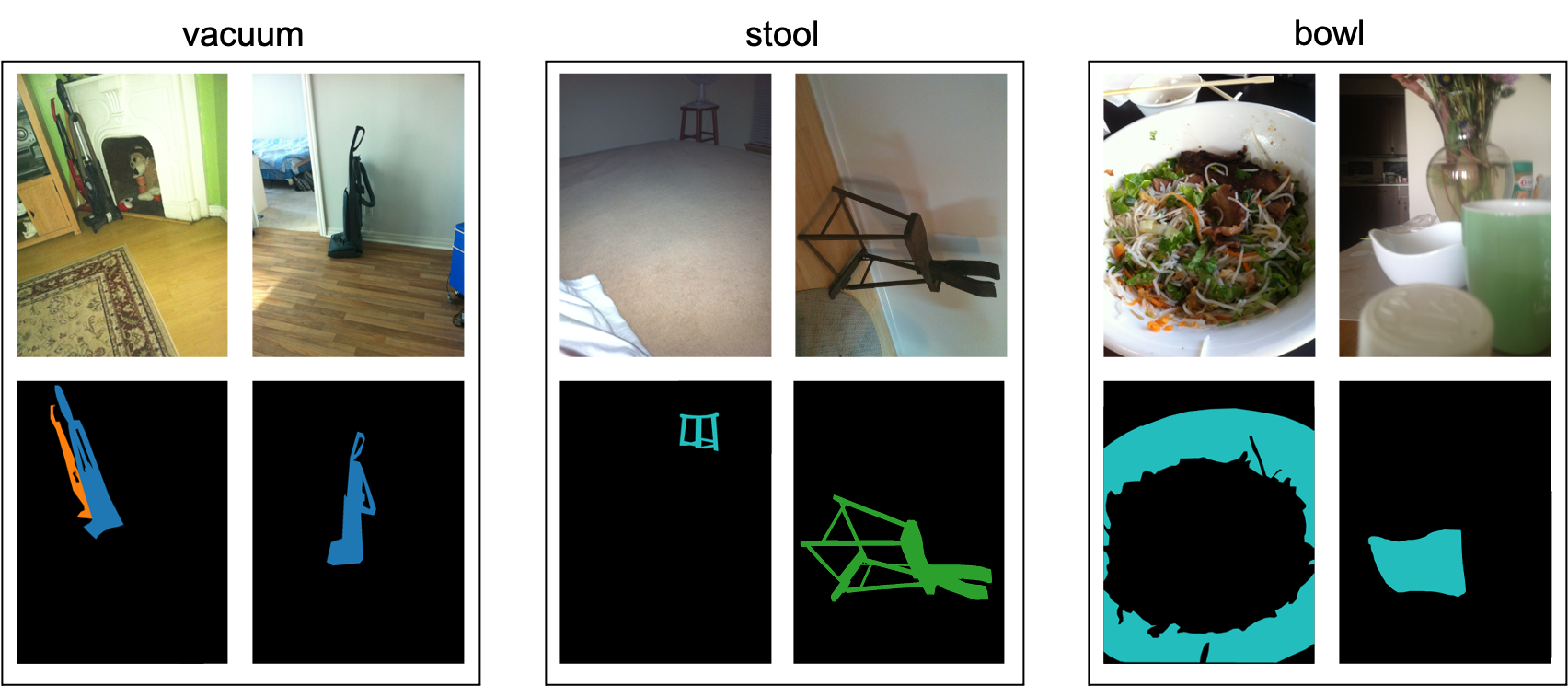}
    \caption{Examples of object categories which frequently contain holes.}
    \label{fig:holes_examples}
\centering
	\includegraphics[width=1.0\textwidth]{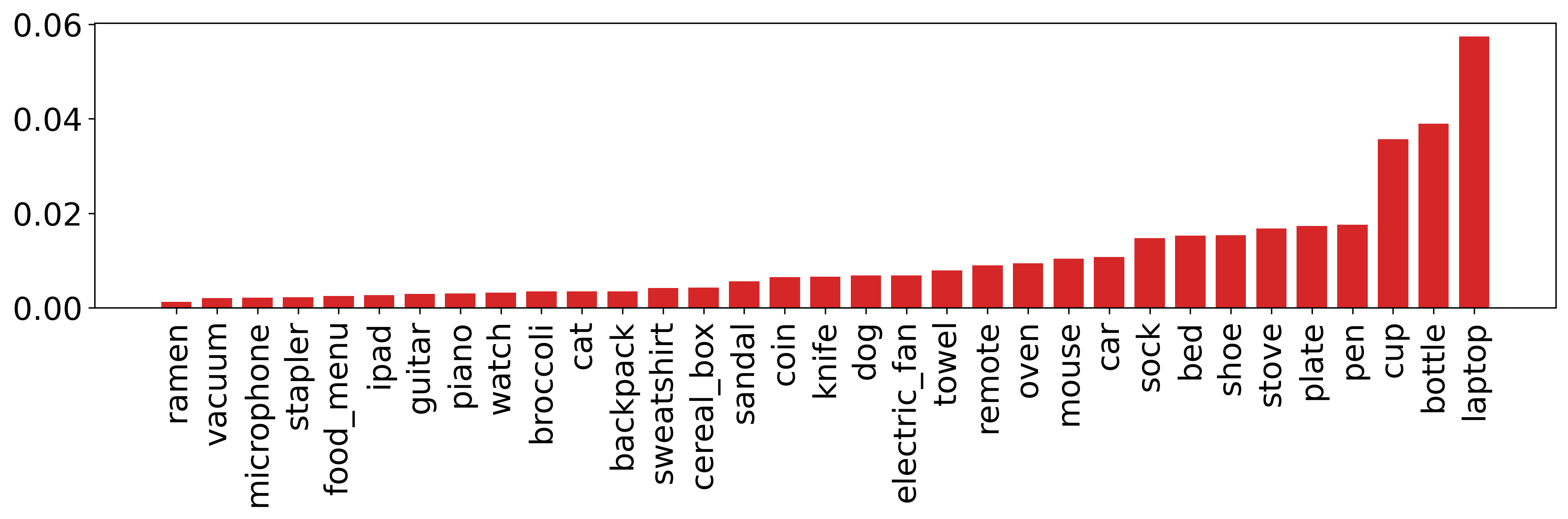} \hfill 
  \vspace{-2em}
  \caption{Proportion of instances corresponding to each category for each third category in our dataset, sorted by frequency of holes.}
  \label{img:categorical_distribution}
\end{figure}

We find that our dataset has, on average, 2.17 annotated objects per image in contrast to 7.33 per image in COCO. We additionally find that each category in our dataset, on average, represents 1.00\% of all instances in comparison to COCO where each category represents, on average, 1.25\% of instances. We visualize a subset of our dataset's categorical distribution in Figure~\ref{img:categorical_distribution}.

We analyze the co-occurences of categories across the images in our dataset. We observe that our dataset has, on average, 2.29 co-occurences per image versus 8.87 in COCO. We find the ten most co-occurring categories, in order, to be keyboard/monitor, person/rug, person/shoe, person/cup, person/bottle, person/sock, person/chair, person/couch, dog/collar, and bed/pillow. We suspect that the person category appears frequently in co-occurrences because many of the common co-occurring categories are worn or held by people.

\section{Examples of few-shot object detection results on VizWiz-FewShot-OD-25\textsuperscript{i}}

\begin{figure}[t!]
    \centering
    \includegraphics[width=0.9\textwidth]{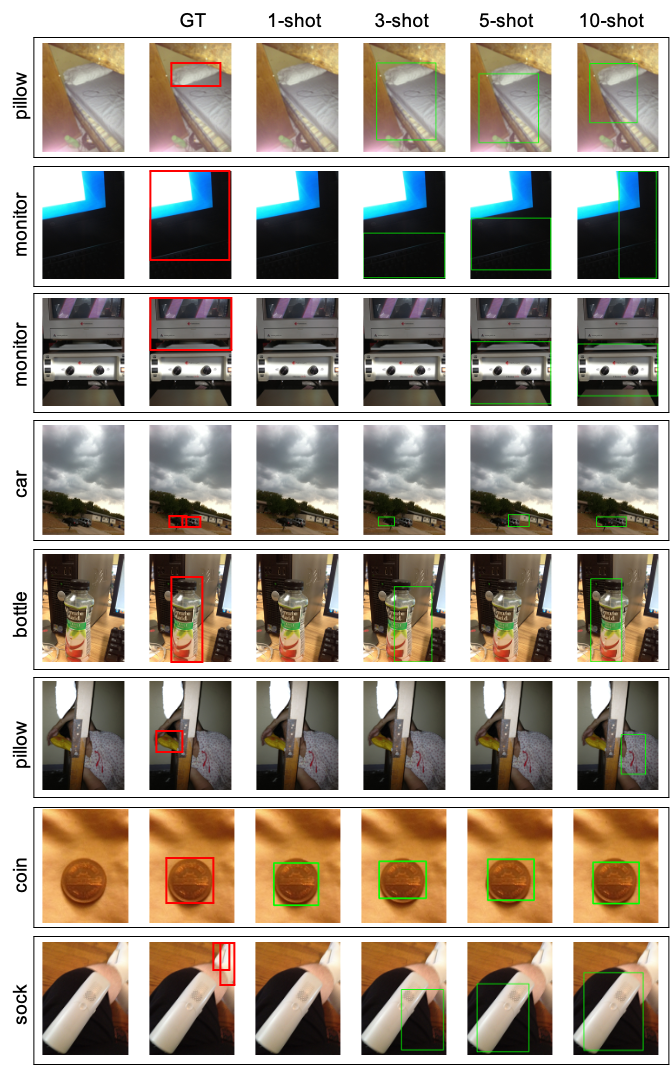}
    \caption{1-, 3-, 5-, and 10-shot object detection results from state-of-the-art few-shot object detection model, DeFRCN~\cite{fsod_DeFRCN}, on our VizWiz-FewShot-OD-25\textsuperscript{i}. While improvements are seen in some examples with more shots, the 10-shot object detection model is not able to perform well on our dataset.}
    \label{fig:fsod_results}
\end{figure}

\begin{figure}[t!]
    \centering
    \includegraphics[width=1\textwidth]{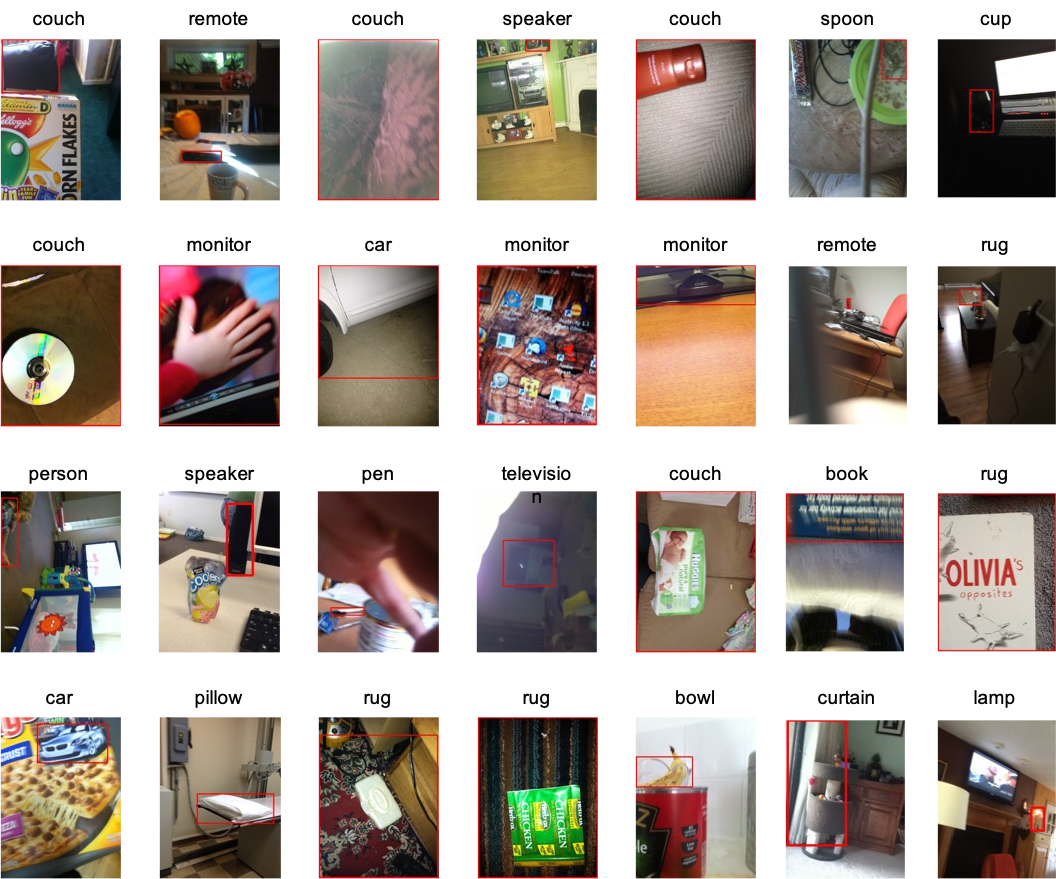}
    \caption{Examples of difficult cases where the state-of-the-art few-shot object detection model, DeFRCN~\cite{fsod_DeFRCN}, is not able to detect the correct object categories. Difficult cases include a large amount of background and partially visible objects.}
    \label{fig:difficult_data}
\end{figure}

We visualize the detection results from the few-shot object detection state-of-the-art model, DeFRCN~\cite{fsod_DeFRCN}.  We randomly selected these images.  For each example, we show one object category and the predictions in that category from 1-, 3-, 5-, and 10-shot models, respectively. Results are shown in Figure~\ref{fig:fsod_results}.  In some cases, none of the 1-, 3-, 5-, and 10-shot models are able to detect the ground truth object category. Examples of these cases are shown in Figure~\ref{fig:difficult_data}.  We observe that these extremely difficult cases include large background objects (i.e. rugs and couches) and objects with low visibility of identifying features (i.e. people and monitors).

\clearpage

\begin{thebibliography}{10}
\providecommand{\url}[1]{\texttt{#1}}
\providecommand{\urlprefix}{URL }
\providecommand{\doi}[1]{https://doi.org/#1}

\bibitem{fsss_oneshot}
Amirreza~Shaban, Shray~Bansal, Z.L.I.E., Boots, B.: One-shot learning for
  semantic segmentation. In: Proceedings of the British Machine Vision
  Conference (BMVC). pp. 167.1--167.13 (September 2017)

\bibitem{bhattacharya2019does}
Bhattacharya, N., Li, Q., Gurari, D.: Why does a visual question have different
  answers? In: Proceedings of the IEEE/CVF International Conference on Computer
  Vision. pp. 4271--4280 (2019)

\bibitem{bigham2010vizwiz}
Bigham, J.P., Jayant, C., Ji, H., Little, G., Miller, A., Miller, R.C., Miller,
  R., Tatarowicz, A., White, B., White, S., et~al.: Vizwiz: nearly real-time
  answers to visual questions. In: Proceedings of the 23nd annual ACM symposium
  on User interface software and technology. pp. 333--342 (2010)

\bibitem{afb_magnifiers}
for~the Blind, A.F.: Low vision optical devices,
  \url{https://www.afb.org/node/16207/low-vision-optical-devices}

\bibitem{fsis_yolact}
Bolya, D., Zhou, C., Xiao, F., Lee, Y.J.: Yolact: {Real-time} instance
  segmentation. In: ICCV (2019)

\bibitem{chen2022grounding}
Chen, C., Anjum, S., Gurari, D.: Grounding answers for visual questions asked
  by visually impaired people. In: Proceedings of the IEEE/CVF Conference on
  Computer Vision and Pattern Recognition. pp. 19098--19107 (2022)

\bibitem{chiu2020assessing}
Chiu, T.Y., Zhao, Y., Gurari, D.: Assessing image quality issues for real-world
  problems. In: Proceedings of the IEEE/CVF Conference on Computer Vision and
  Pattern Recognition. pp. 3646--3656 (2020)

\bibitem{dataset_imagenet}
Deng, J., Dong, W., Socher, R., Li, L.J., Li, K., Fei-Fei, L.: Imagenet: A
  large-scale hierarchical image database. In: 2009 IEEE Conference on Computer
  Vision and Pattern Recognition. pp. 248--255 (2009).
  \doi{10.1109/CVPR.2009.5206848}

\bibitem{desmond_microsofts_nodate}
Desmond, N.: Microsoft’s {Seeing} {AI} founder {Saqib} {Shaikh} is speaking
  at {Sight} {Tech} {Global},
  \url{https://social.techcrunch.com/2020/08/20/microsofts-seeing-ai-founder-saqib-shaikh-is-speaking-at-sight-tech-global/}

\bibitem{FSOD_fewExample}
Dong, X., Zheng, L., Ma, F., Yang, Y., Meng, D.: Few-example object detection
  with model communication. IEEE Transactions on Pattern Analysis and Machine
  Intelligence  \textbf{PP}, ~1--1 (06 2018)

\bibitem{dataset_voc}
Everingham, M., Gool, L.V., Williams, C.K.I., Winn, J.M., Zisserman, A.: The
  pascal visual object classes (voc) challenge. International Journal of
  Computer Vision (IJCV)  \textbf{88},  303--338 (2009)

\bibitem{be_my_eyes_be_nodate}
Eyes, B.M.: Be {My} {Eyes}: {Our} story, \url{https://www.bemyeyes.com/about}

\bibitem{fsod_attenRPN}
Fan, Q., Zhuo, W., Tang, C.K., Tai, Y.W.: Few-shot object detection with
  attention-rpn and multi-relation detector. In: Proceedings of the IEEE/CVF
  Conference on Computer Vision and Pattern Recognition (CVPR) (2020)

\bibitem{fsis_FGN}
Fan, Z., Yu, J., Liang, Z., Ou, J., Gao, C., Xia, G., Li, Y.: {FGN:} fully
  guided network for few-shot instance segmentation. In: 2020 {IEEE/CVF}
  Conference on Computer Vision and Pattern Recognition (CVPR). pp. 9169--9178.
  Computer Vision Foundation / {IEEE} (2020)

\bibitem{gurari2018predicting}
Gurari, D., He, K., Xiong, B., Zhang, J., Sameki, M., Jain, S.D., Sclaroff, S.,
  Betke, M., Grauman, K.: Predicting foreground object ambiguity and
  efficiently crowdsourcing the segmentation (s). International Journal of
  Computer Vision  \textbf{126}(7),  714--730 (2018)

\bibitem{gurari2019vizwiz}
Gurari, D., Li, Q., Lin, C., Zhao, Y., Guo, A., Stangl, A., Bigham, J.P.:
  Vizwiz-priv: A dataset for recognizing the presence and purpose of private
  visual information in images taken by blind people. In: Proceedings of the
  IEEE/CVF Conference on Computer Vision and Pattern Recognition. pp. 939--948
  (2019)

\bibitem{dataset_VizWiz}
Gurari, D., Li, Q., Stangl, A.J., Guo, A., Lin, C., Grauman, K., Luo, J.,
  Bigham, J.P.: Vizwiz grand challenge: Answering visual questions from blind
  people. In: Proceedings of the IEEE Conference on Computer Vision and Pattern
  Recognition. pp. 3608--3617 (2018)

\bibitem{dataset_VizWizCaption}
Gurari, D., Zhao, Y., Zhang, M., Bhattacharya, N.: Captioning images taken by
  people who are blind. In: ECCV (2020)

\bibitem{dataset_KVAQ}
J.-H.~Kim, S.~Lim, J.P.H.C.: Korean localization of visual question answering
  for blind people. SK T-Brain - AI for Social Good Workshop at NeurIPS  (2019)

\bibitem{fsod_review1}
Jiaxu, L., Taiyue, C., Xinbo, G., Yongtao, Y., Ye, W., Feng, G., Yue, W.: A
  comparative review of recent few-shot object detection algorithms (2021)

\bibitem{fsod_FSRW}
Kang, B., Liu, Z., Wang, X., Yu, F., Feng, J., Darrell, T.: Few-shot object
  detection via feature reweighting. In: 2019 IEEE/CVF International Conference
  on Computer Vision (ICCV). pp. 8419--8428 (Nov 2019)

\bibitem{lee2020emerging}
Lee, S., Reddie, M., Tsai, C.H., Beck, J., Rosson, M.B., Carroll, J.M.: The
  emerging professional practice of remote sighted assistance for people with
  visual impairments. In: Proceedings of the 2020 CHI Conference on Human
  Factors in Computing Systems. pp. 1--12 (2020)

\bibitem{dataset_FSS1000}
Li, X., Wei, T., Chen, Y.P., Tai, Y.W., Tang, C.K.: Fss-1000: A 1000-class
  dataset for few-shot segmentation. Proceedings of the IEEE/CVF Conference on
  Computer Vision and Pattern Recognition (CVPR)  (2020)

\bibitem{dataset_coco}
Lin, T.Y., Maire, M., Belongie, S., Hays, J., Perona, P., Ramanan, D.,
  Doll{\'a}r, P., Zitnick, C.L.: Microsoft coco: Common objects in context. In:
  European Conference on Computer Vision (ECCV). pp. 740--755 (2014)

\bibitem{dataset_orbit}
Massiceti, D., Zintgraf, L., Bronskill, J., Theodorou, L., Tobias~Harris, M.,
  Cutrell, E., Morrison, C., Hofmann, K., Stumpf, S.: Orbit: A real-world
  few-shot dataset for teachable object recognition. In: ICCV 2021 (October
  2021)

\bibitem{fsis_siameseMRCNN}
Michaelis, C., Ustyuzhaninov, I., Bethge, M., Ecker, A.S.: One-shot instance
  segmentation. ArXiv  (2018)

\bibitem{fsis_FAPIS}
Nguyen, K., Todorovic, S.: Fapis: A few-shot anchor-free part-based instance
  segmenter. 2021 IEEE/CVF Conference on Computer Vision and Pattern
  Recognition (CVPR) pp. 11094--11103 (2021)

\bibitem{fsss_FeatureWeighting}
Nguyen, K.D.M., Todorovic, S.: Feature weighting and boosting for few-shot
  segmentation. 2019 IEEE/CVF International Conference on Computer Vision
  (ICCV) pp. 622--631 (2019)

\bibitem{fsod_DeFRCN}
Qiao, L., Zhao, Y., Li, Z., Qiu, X., Wu, J., Zhang, C.: Defrcn: Decoupled
  faster r-cnn for few-shot object detection. ArXiv  (2021)

\bibitem{stangl2018browsewithme}
Stangl, A.J., Kothari, E., Jain, S.D., Yeh, T., Grauman, K., Gurari, D.:
  Browsewithme: An online clothes shopping assistant for people with visual
  impairments. In: Proceedings of the 20th International ACM SIGACCESS
  Conference on Computers and Accessibility. pp. 107--118 (2018)

\bibitem{fsod_metarcnn}
Yan, X., Chen, Z., Xu, A., Wang, X., Liang, X., Lin, L.: Meta r-cnn: Towards
  general solver for instance-level low-shot learning. 2019 IEEE/CVF
  International Conference on Computer Vision (ICCV)  (Oct 2019)

\bibitem{zeng2020vision}
Zeng, X., Wang, Y., Chiu, T.Y., Bhattacharya, N., Gurari, D.: Vision skills
  needed to answer visual questions. Proceedings of the ACM on Human-Computer
  Interaction  \textbf{4}(CSCW2),  1--31 (2020)

\bibitem{OD_review}
Zhao, Z.Q., Zheng, P., Xu, S.T., Wu, X.: Object detection with deep learning: A
  review. IEEE Transactions on Neural Networks and Learning Systems
  \textbf{PP},  1--21 (01 2019). \doi{10.1109/TNNLS.2018.2876865}

\end{thebibliography}
\end{document}